\documentclass[sigconf]{acmart}

\AtBeginDocument{%
  \providecommand\BibTeX{{%
    \normalfont B\kern-0.5em{\scshape i\kern-0.25em b}\kern-0.8em\TeX}}}

\setcopyright{none}
\renewcommand\footnotetextcopyrightpermission[1]{} 

\usepackage{amsmath}

\usepackage{amssymb}
\usepackage{mathtools}
\usepackage{amsthm}

\usepackage[capitalize,noabbrev]{cleveref}

\theoremstyle{plain}
\newtheorem{assumption}{Assumption}
\newtheorem{theorem}{Theorem}[section]
\newtheorem{proposition}[theorem]{Proposition}
\newtheorem{conjecture}[theorem]{Conjecture}
\newtheorem{lemma}[theorem]{Lemma}
\newtheorem{corollary}[theorem]{Corollary}
\theoremstyle{definition}
\newtheorem{definition}[theorem]{Definition}
\usepackage[textsize=tiny]{todonotes}

\usepackage{amsmath,amsfonts,bm}

\makeatletter
\def\widebreve{\mathpalette\wide@breve}
\def\wide@breve#1#2{\sbox\z@{$#1#2$}%
     \mathop{\vbox{\m@th\ialign{##\crcr
\kern0.08em\brevefill#1{0.6\wd\z@}\crcr\noalign{\nointerlineskip}%
                    $\hss#1#2\hss$\crcr}}}\limits}
\def\brevefill#1#2{$\m@th\sbox\tw@{$#1($}%
  \hss\resizebox{#2}{\wd\tw@}{\rotatebox[origin=c]{90}{\upshape(}}\hss$}
\makeatletter

\def\1{\bm{1}}

\DeclareMathAlphabet{\mathsfit}{\encodingdefault}{\sfdefault}{m}{sl}
\SetMathAlphabet{\mathsfit}{bold}{\encodingdefault}{\sfdefault}{bx}{n}

\newcommand{\iid}{\ensuremath{\text{i.i.d.}}\xspace}

\newcommand{\noniid}{\ensuremath{\text{non-i.i.d.}}\xspace}

\newcommand{\hessian}{\mathcal{H}}

\usepackage{todonotes}
\usepackage{xspace}
\usepackage{multirow}

\usepackage{enumitem}

\newcommand{\para}[1]{\paragraph{\textbf{#1}}}
\newcommand{\ours}{FedMeZO\xspace}
\newcommand{\ourtopic}{ZOO-FL\xspace}

\newcommand{\ie}{\textit{i.e.}\xspace}

\begin{document}

\pagestyle{plain}

\title[On the Convergence of Zeroth-Order Federated Tuning for Large Language Models]{On the Convergence of Zeroth-Order Federated Tuning for \\ Large Language Models}



\author{Zhenqing Ling}
\affiliation{%
  \institution{Sun Yat-sen University}
  \city{Shenzhen}
  \state{Guangdong}
  \country{China}
}
\email{lingzhq@mail2.sysu.edu.cn}

\author{Daoyuan Chen}
\affiliation{%
  \institution{Alibaba Group}
  \city{Hangzhou}
  \state{Zhejiang}
  \country{China}
}
\email{daoyuanchen.cdy@alibaba-inc.com}

\author{Liuyi Yao}
\affiliation{%
  \institution{Alibaba Group}
  \city{Hangzhou}
  \state{Zhejiang}
  \country{China}
}
\email{yly287738@alibaba-inc.com}

\author{Yaliang Li}
\affiliation{%
  \institution{Alibaba Group}
  \city{Bellevue}
  \state{Washington}
  \country{United States}
}
\email{yaliang.li@alibaba-inc.com}

\author{Ying Shen}
\authornote{Corresponding author.}
\affiliation{%
  \institution{Sun Yat-sen University \\}
  \city{Shenzhen}
  \state{Guangdong}
  \country{China\\}
  \institution{Pazhou Lab}
  \city{Guangzhou}
  \state{Guangdong}
  \country{China}
  \institution{Guangdong Provincial Key Laboratory of Fire Science and Intelligent Emergency Technology}
  \city{Guangzhou}
  \state{Guangdong}
  \country{China}
}
\email{sheny76@mail.sysu.edu.cn}



\begin{abstract}
The confluence of Federated Learning (FL) and Large Language Models (LLMs) is ushering in a new era in privacy-preserving natural language processing. However, the intensive memory requirements for fine-tuning LLMs pose significant challenges, especially when deploying on clients with limited computational resources. To circumvent this, we explore the novel integration of Memory-efficient Zeroth-Order Optimization within a federated setting, a synergy we term as FedMeZO. 
Our study is the first to examine the theoretical underpinnings of FedMeZO in the context of LLMs, tackling key questions regarding the influence of large parameter spaces on optimization behavior, the establishment of convergence properties, and the identification of critical parameters for convergence to inform personalized federated strategies. 
Our extensive empirical evidence supports the theory, showing that FedMeZO not only converges faster than traditional first-order methods such as FedAvg but also significantly reduces GPU memory usage during training to levels comparable to those during inference. 
Moreover, the proposed personalized FL strategy that is built upon the theoretical insights to customize the client-wise learning rate can effectively accelerate loss reduction. 
We hope our work can help to bridge theoretical and practical aspects of federated fine-tuning for LLMs, thereby stimulating further advancements and research in this area.
\end{abstract}

\maketitle

\footnotetext{© {Zhenqing Ling et al. | ACM} {2024}. This is the author's version of the work. It is posted here for your personal use. Not for redistribution. The definitive Version of Record was published in SIGKDD 2024, http://dx.doi.org/10.1145/3637528.3671865.}

\section{Introduction}
\label{sec:intro}

Federated Learning (FL) has become an important approach in modern machine learning, particularly in scenarios where data decentralization and privacy-preserving are crucial \citep{muhammad2020fedfast,meng2021cross,hong2021federated,fsgnn,fs}. Central to this learning paradigm is the corroborative training of a global model through the aggregation of updates from multiple clients, without sharing their raw data \citep{mcmahan2017communication,kairouz2019advances}.

In parallel, Large Language Models (LLMs) have radically advanced the field of natural language processing \citep{touvron2023llama, du2021glm, brown2020language}. 
The fine-tuning of LLMs that are already pre-trained on vast corpora, has proven to be a highly effective strategy for numerous tasks, yielding models that are both versatile and capable of adapting to specific domain narratives or aligning with human values \cite{malladi2023fine,brown2020language}.

The tuning of LLMs requires suitable alignment data, which are often costly to acquire \cite{DatabricksBlog2023DollyV2,chen2023datajuicer}. Due to the abundance of private data that remains largely isolated and underutilized, the intersection of FL and LLMs has sparked increasing interest among researchers \cite{zhang2023fedit,kuang2023federatedscope-LLM,fan2023-FATE-LLM,qin2024federated}. Notably, this integration presents significant computational challenges, especially for clients with limited resources \cite{zhang2023towards,chen2023fsreal}. The scaling up of LLMs further compounds this issue, as the computation of gradients for backpropagation incurs substantial memory costs, frequently surpassing the practical capabilities of these clients \cite{malladi2023fine}.

Addressing this challenge, we turn our attention to Zeroth-Order Optimization (ZOO), an algorithm that computes gradient approximations without explicit gradient information, thus significantly reducing memory consumption \cite{gu2021efficient}. However, the combination of ZOO and FL —a research direction we refer to as \ourtopic—remains unexplored in the literature in the context of LLMs \cite{zerofl}. Our work intends to bridge this gap by harnessing the memory efficiency of ZOO within the context of federated fine-tuning of LLMs, especially on the following theoretical foundations:

(\textbf{Q1}) How does the vast parameter space of LLMs influence the behavior of \ourtopic? (\textbf{Q2}) Can we establish the convergence properties of \ourtopic for LLMs? (\textbf{Q3}) Which model parameters are critical for convergence, and how can we leverage them to optimize FL performance, such as via personalization?

In this paper, we focus on incorporating a memory-efficient ZOO method, MeZO \cite{malladi2023fine} into FL, a synergy we denote as \ours, and establishing its convergence properties under the large-scale parameter space of LLMs. We analyze and present precise convergence rates characterized by the low effective rank $r$ of the models' Hessian matrices \citep{aghajanyan2021intrinsic, li2018measuring}, and other typical FL parameters such as number of clients $N$, number of FL rounds $T$, iteration steps of local training $H$ and heterogeneity constants $c_h$ and $\sigma_h$.
Moreover, we reveal the learning rate to be a crucial variable for convergence.
Building on theoretical insights, we further propose a strategy that tailors the learning rate to each client's specific data characteristics. 

To validate our theoretical results, we conduct extensive experiments on real-world FL datasets for LLM tuning, which cover diverse data distributions and application tasks.
Our empirical findings corroborate the theoretical analysis, validating effective convergence even when scaling up to models with billions of parameters. 
Compared with first-order methods such as FedAvg, \ours converges faster meanwhile remarkably reducing GPU memory requirements. 
The personalized strategies guided by our theoretical insights, empirically show a more rapid loss reduction, as opposed to non-personalized or random learning rate assignments.

In summary, our theoretical and empirical exploration validates \ours in the fine-tuning process of LLMs, providing a rigorous framework and practical insights for future applications. Our key contributions are threefold:
\begin{itemize}[leftmargin=*]
\item We advance the understanding of \ours for LLMs, extending the two-point gradient estimation to federated tuning and establishing theoretical convergence rate $\mathcal{O}\left( r^{3/2} (NHT)^{-1/2} \right)$ and $\mathcal{O}\left( r^{3/2} (\widetilde{c}_h NHT)^{-1/2} \right)$ $-\mathcal{O}\Bigl( \sigma_h^2 (c_h N)^{-1} \Bigr)$ for the \iid setting and \noniid setting, respectively.
\item We analyze the impact of various hyperparameters of \ours and explore a theory-informed strategy for personalized learning rate adjustment, providing practical guidance for \ourtopic.
\item Through extensive empirical evidence with LLMs, we verify the proposed theoretical results and show that \ours yields effective convergence with substantially reduced memory overhead compared to FedAvg. 
Our codes are publicly available at 
\url{https://github.com/alibaba/FederatedScope/tree/FedMeZO}.
\end{itemize}




\section{Preliminaries} 

\subsection{Background and Related Works}
\label{sec:related-work}

\para{Federated Fine-Tuning of Large Language Models.}
Large Language Models (LLMs) have demonstrated remarkable capabilities that enable a variety of real-world applications \cite{zhang2022opt,touvron2023llama,zhao2023survey}. The federated fine-tuning of LLMs has recently attracted attention, focused on adapting these models to domain-specific tasks while preserving the privacy of the training data. Chen et al. \cite{chen2023federated} investigated the integration of LLMs within federated settings, highlighting the inherent challenges and potential opportunities. Zhang et al. \cite{zhang2023towards} furthered this research by examining instruction tuning of LLMs in a federated context, marking progress in applying FL to the specialized training of LLMs. Notable frameworks such as FATE-LLM by Fan et al. \cite{fan2023fate} and FederatedScope-LLM by Kuang et al. \cite{kuang2023federatedscope-LLM} offer industrial-grade and comprehensive solutions for federated fine-tuning. Our work, in contrast, investigates the fusion of Zeroth-Order Optimization (ZOO) with FL for the fine-tuning of LLMs, an area that has yet to be fully investigated, thereby addressing a gap in the literature and providing fundamental theoretical insights.

\para{Zeroth-Order Optimization in Federated Learning.}
ZOO has emerged as a viable method to address the difficulties of computing gradients in FL, especially in settings limited by computational resources. Zhang et al. \cite{zhang2021desirable} proposed a ZOO algorithm tailored for vertical FL, focusing on privacy preservation. Yi et al. \cite{yi2022zeroth} and Li et al. \cite{li2021communication} studied ZOO-FL algorithms, with discussions on convergence properties with single-point perturbation and local updates in decentralized FL, respectively. 
The convergence analysis is a critical aspect of FL, as illustrated by Li et al. \cite{li2019convergence} for the FedAvg algorithm and further developed by Fang et al. \cite{fang2022communication} for mini-batch stochastic ZOO-FL in wireless networks. Moreover, Shu et al. \cite{shu2023federated} proposed enhancements to query efficiency for ZOO within the FL framework. 
Our research sets itself apart by formulating theoretical convergence bounds for ZOO-FL, specifically tailored to the large-scale parameter space of LLMs.
This builds on the preliminary work by Malladi et al. \cite{malladi2023fine}, which confirmed the feasibility of ZOO for LLMs in a centralized setting.

\subsection{Problem Formulation}
\label{subsec:problem_formula}
For readability and brevity, we summarize the full list of introduced notations in Appendix \ref{appendix:notation} and present detailed proofs of all theoretical results in Appendix \ref{appendix:assumptions}-\ref{appendix:theorem}.

\para{Federated Learning.}
We consider the general FL setting as of FedAvg \cite{mcmahan2017communication}, with a central server and a collection of $N$ clients, indexed by ${1, 2, ..., N}$. 
The central server coordinates the training of a global model through the collaborative efforts of these clients, each holding local data samples drawn from their respective distributions $\mathcal{D}i$. The optimization problem can be formulated as:
\begin{equation}
\label{eq:pre_fl_origin}
    \min_{\theta \in \mathbb{R}^d} f(\theta) \stackrel{\triangle}{=} \frac{1}{N} \sum_{i=1}^N f_i(\theta_i), \quad f_i(\theta) \stackrel{\triangle}{=} \mathbb{E}_{\mathcal{B}_i \sim \mathcal{D}_i}\bigl[ F_i(\theta, \mathcal{B}_i) \bigr],
\end{equation}
where $\theta \in \mathbb{R}^d$ denotes the $d$-dimension parameter of the model, and $f(\theta)$ and $f_i(\theta)$ denote the global loss function on the central server and local loss function on $i^{th}$ client, respectively. 
Typically, the clients are assumed with equal importance \cite{wang2021field}, and the data is randomly sampled for efficiency \cite{li2014efficient}. $F_i(\theta, \mathcal{B}_i)$ represents the local loss function w.r.t a specific mini-batch $\mathcal{B}_i$ drawn from $\mathcal{D}_i$.
 
\para{Zeroth-Order Optimization.}

Zeroth-order optimization (ZOO) is a prominent technique in scenarios where gradients are difficult to obtain, which estimates gradients by forward propagations. Given a random vector $z$ and a smoothing constant $\mu$, a typical one-point gradient estimator \cite{duchi2015optimal} is defined as:
\begin{align}
\label{eq:pre_one_point_estimator_def}
\widetilde{\nabla} F(\theta, z, \mathcal{B}, \mu) = \frac{z}{2\mu} \bigl(F(\theta+\mu z, \mathcal{B}) - F(\theta, \mathcal{B})\bigr),
\end{align}

However, Eq. \eqref{eq:pre_one_point_estimator_def} provides a biased gradient estimation, leading to a certain degree of information loss \cite{liu2020primer}. Hence our work employs the ZOO paradigm with a two-point gradient estimator proposed by \cite{malladi2023fine} in a federated setting:
\begin{definition}
\label{def:two_point_zoo}
(Two-point gradient estimator) Given a set of parameters $\theta \in \mathbb{R}^d$ for an LLM and a mini-batch $\mathcal{B}_i$, the two-point zeroth-order gradient estimator is formulated as:
\begin{align}
\label{eq:pre_two_point_estimator_def}
\widetilde{\nabla} F_i(\theta, z_i, \mathcal{B}_i, \mu) = \frac{z_i}{2\mu} \bigl(F_i(\theta+\mu z_i, \mathcal{B}_i) - F_i(\theta-\mu z_i, \mathcal{B}_i)\bigr),
\end{align}
where $z_i \sim \mathcal{N}(0, I_d)$ is a Gaussian random variable and $\mu$ is the perturbation scale.
The two-point gradient estimator in Eq. \eqref{eq:pre_two_point_estimator_def} requires only two forward passes through the model to compute the estimation of gradient, which serves as a memory-efficient alternative to backpropagation (BP).
\end{definition}

\para{The \ours Algorithm. }
In this paper, we study and analyze the proprieties of a practical synergy of MeZO \cite{malladi2023fine} and FedAvg \cite{mcmahan2017communication}, which is designed to fine-tune LLMs in an efficient, privacy-preserving and personalized manner.
We term this \ourtopic approach as \ours, depicted with the following processes:

In a single communication round, the central server first broadcasts the global model parameters to available clients. Once the clients have completed their local updates and uploaded their models, the server aggregates the updates according to Eq. \eqref{eq:pre_fl_origin}, forming the basis for the subsequent round.

Upon receiving the global model parameters, clients perform the following steps, distinguishing \ours from traditional BP-based FedAvg algorithms in two-fold:

    \textit{(1) Training Memory Reduction:} Clients update their models using the two-point ZOO gradient estimator defined in Eq. \eqref{eq:pre_two_point_estimator_def} as:
        \begin{equation}
    \label{eq:pre_e_i}
        e_i^{(t, k)} = \widetilde{\nabla}{F_i(\theta_i^{(t,k)}, z_i^{(t,k)}, \mathcal{B}_i^{(t,k)}, \mu)},
    \end{equation}
    where $(t,k)$ denotes the $k^{th}$ iteration within the $t^{th}$ communication round. Unlike standard ZO-SGD algorithms that require storing the perturbation vector $z$ at each iteration, \ours resamples $z$ using random seeds in in-place implementation, thus reducing memory usage to a level equivalent to inference \cite{malladi2023fine}.
    
    \textit{(2) Communication Cost Reduction:} To mitigate the high communication overhead associated with LLMs, \ours leverages Low-Rank Adaptation (LoRA) \cite{hu2022lora,bai2024federated}, which introduces reparametrization to tune two small delta matrix on the linear layers instead of the whole LLM weights, based on the assumption that well pre-trained LLM possess a low ``intrinsic dimension'' when adapted to new tasks. Introducing it can help us further reduce the number of parameters to be updated and uploaded, thereby aligning with the practical constraints of federated settings. Detailed analysis of communication cost is available in Appendix \ref{appendix:communication_cost}.

\subsection{Lemmas and Assumptions}

\begin{lemma}
\label{lemma:unbiased}
\textbf{(Unbiased Gradient Estimator)} The two-point zeroth-order gradient estimator described in Eq. \eqref{eq:pre_two_point_estimator_def} is an unbiased estimator of the true gradient, that is,
\begin{equation}
\label{eq:pre_unbiased}
\mathbb{E}[ \widetilde{\nabla} F_i(\theta, z_i, \mathcal{B}_i, \mu) ] = \nabla f_i(\theta).
\end{equation}
\end{lemma}

The Hessian matrix, which is the square matrix of second-order partial derivatives of the loss w.r.t the model parameters, characterizes the curvature of the loss surface \cite{ghorbani2019investigation}. Although the size of a model's loss Hessian is often associated with the rate of fine-tuning, studies suggest that the large-scale parameters of LLMs do not necessarily impede convergence \citep{jamieson2012query, agarwal2009information}. 
This paradox is addressed by recognizing that the loss Hessian often exhibits a small local effective rank \cite{malladi2023fine}, which we capture in the following assumption:

\begin{assumption}
\label{assumption:low_effective_rank}
There exist a Hessian matrix $\mathcal{H}(\theta^t)$ satisfying:

$\bullet$ $\nabla^2 f(\theta) \preceq \mathcal{H}(\theta^t)$ for all $\theta$ such that $\lVert \theta-\theta^t \rVert \leq \eta d G(\theta^t)$, where $G(\theta^t)=\max_{\mathcal{B} \sim \mathcal{D}}\lVert \nabla f(\theta^t,\mathcal{B}) \rVert$.

$\bullet$The effective rank of $\mathcal{H}(\theta^t)$, denoted as $\text{tr}(\mathcal{H}(\theta^t)) / \lVert \mathcal{H}(\theta^t) \rVert_{\text{op}}$, is at most $r$. Here $\text{tr}$ denotes the trace of the matrix, and $\lVert \cdot \rVert_{\text{op}}$ denotes the operator norm.
\end{assumption}

Assumption \ref{assumption:low_effective_rank} characterizes a low effective rank $r$ in the Hessian matrix, which demonstrates that LLM fine-tuning can occur in a low dimensional subspace ($\leq 200$ parameters) \citep{aghajanyan2021intrinsic, li2018measuring}.
With this insight, \cite{malladi2023fine} identified the bound of loss descent at each step of centralized ZOO, which is partially influenced by $r$:

\begin{lemma}
\label{lemma:dimension_free}
    \textbf{(Bounded Centralized Descent)} 
    Assume $f(\theta)$ is $L$-smooth and let $\widetilde{\nabla}{F(\theta, z, \mathcal{B}, \mu)}$ be the unbiased zeroth-order gradient estimator from Eq. \eqref{eq:pre_two_point_estimator_def}. If the Hessian matrix $\hessian(\theta)$ exhibits a local effective rank of $r$, and constants $\gamma = \Theta(r / n)$ and $\zeta = \Theta(1 / r d)$ exist, then the expected decrease in loss can be bounded as follows:
    \begin{align}
    \label{eq:pre_each_step_low_rank_estimator}
        \mathbb{E}\bigl[ f(\theta^{t+1}) \bigr] \leq f(\theta^{t}) &-  \frac{\eta}{\gamma} {\lVert \nabla f(\theta^{t}) \rVert}^2 
        \notag \\
        &+ \frac{\eta^2 L \zeta}{2}    \mathbb{E}\bigl[ {\lVert  \widetilde{\nabla}{F(\theta, z, \mathcal{B}, \mu)} \rVert}^2 \bigr], \
    \end{align}
\end{lemma}
\noindent where $\gamma = \frac{dr +d-2}{n(d+2)}$, $\zeta = \frac{(d+2)n^2}{(dr+d-2)(d+n-1)}$, and $n$ is the number of randomizations.

From Eq. \eqref{eq:pre_each_step_low_rank_estimator}, we observe that the rate of descent at a single step depends on the gradient related to $\gamma$ and the gradient estimation related to $\zeta$. Following  \cite{malladi2023fine}, we set $n$ to 1 in this paper.

Besides, to facilitate the analysis in FL setting, we introduce four assumptions, including \textit{Bounded Loss} (Assumption \ref{assumption:lower_bound}), \textit{$L$-smoothness} (Assumption \ref{assumption:lsmooth}), \textit{mini-batch gradient error bound} (Assumption \ref{assumption:minibatch}, \textit{global-local disparities in \iid and \noniid settings} (Assumptions \ref{assumption:local_global_iid} and \ref{assumption:local_global_noniid} respectively).
These assumptions are standard and foundational in optimization and FL literature \cite{bottou2018optimization,li2019convergence, wang2021novel, li2020federated}, which we detail in Appendix \ref{appendix:assumptions}.

\section{Main results}
\label{sec:main_results}

\subsection{Convergence Analysis in \iid Case}
\label{subsec:result_iid_convergence}
In this subsection, we examine the convergence properties of \ours within the \iid data distribution setting. We establish the conditions under which the algorithm guarantees loss reduction at each iteration and provide a global convergence rate.

\begin{theorem}
\label{thm:iid}
\textbf{(Stepwise Loss Descent in \iid Setting)} 
Under Assumptions \ref{assumption:low_effective_rank}-\ref{assumption:local_global_iid} and with a learning rate $\eta$ satisfying
\begin{equation}
\label{eq:eta_satisfies_thm_iid}
        \eta \leq \min \Bigl\{ \frac{1}{3 H L \sqrt{c_g d}}, \frac{N}{3 H L c_g}, \frac{1}{H^2}\Bigr\},
\end{equation}
the expected decrease in loss at each step for \ours under the \iid scenario is bounded as
\begin{align}
\label{eq:each_step_satisfies_thm_iid}
\mathbb{E}_t \left[ f(\theta^{t+1}) \right] &\leq f(\theta^{t}) - \left( \frac{2}{\gamma} - \frac{2 \zeta}{d} \right) \eta \mathbb{E}_t \left\lVert \nabla f(\theta^{t}) \right\rVert^2\notag \\
& \quad + \frac{2 \sigma_g^2 \zeta \eta L}{NHd} + \frac{\zeta \eta \mu^2 L^3 }{2NH},
\end{align}
where $\gamma$ and $\zeta$ quantify the effective low-rank properties of the gradient and its estimator, respectively.
\end{theorem}

In Theorem \ref{thm:iid}, the term $(-2\eta/\gamma) \mathbb{E}_t | \nabla f(\theta^{t}) |^2$ serves as a critical factor that drives the decrease in the loss function, as it is the sole negative contributor in Eq. \eqref{eq:each_step_satisfies_thm_iid} such that $\mathbb{E}_t \bigl[ f(\theta^{t+1})-f(\theta^{t}) \bigr) \bigr] \leq 0$.
Note that the presence of the factor $\gamma^{-1} = \Theta(r^{-1})$ underscores the impact of the low effective rank $r$ on the convergence rate (under Assumption \ref{assumption:low_effective_rank}), revealing that a reduction in $r$ can accelerate convergence independently of the high-dimensional parameter space $d$. Consequently, even for LLMs with expansive parameter spaces, \ours can attain convergence. 
This addresses our first foundational question ``\textit{\textbf{Q1}: How does the vast parameter space of LLMs influence the behavior of \ourtopic?}''.

Moreover, the terms $(2\eta \zeta / d) \mathbb{E}_t \big\lVert \nabla f(\theta^{t}) \big\rVert^2$ and $(2 \sigma_g^2 \zeta \eta L / NHd) $ are scaled by $\zeta/d$, \ie, in $\Theta(1 / rd^2)$, contributing a relatively smaller effect on the convergence speed compared to negative term.
This demonstrates that the influence on convergence speed from the zeroth-order gradient estimation is moderated by the model's effective low rank and dimensionality.
As for the last term, $(\zeta \eta \mu^2 L^3 / 2NH)$, it acts as a factor slowing down the convergence rate, and we can observe that when $N$ and $H$ are larger, this term becomes smaller. This suggests that the effect of slowing down the convergence rate is not as pronounced, and simultaneously, the perturbation step $ \mu$ should not be excessively large. Specifically, this term indicates that increasing the number of clients and the number of local rounds can enhance convergence, while also emphasizing the importance of keeping the perturbation step $ \mu$ moderate.

After gaining intuitive insights in each round of training through the analysis of Theorem \ref{thm:iid}, it is necessary to assess the convergence performance of \ours from a global perspective. 
We utilize the squared magnitude of the gradient $\mathbb{E}_t \lVert \nabla f(\theta^{t}) \rVert^2$ as a measure to assess the suboptimality of each iterate. 
The rapidity with which the algorithm approaches a stationary point serves as a crucial metric for determining its efficacy in the context of non-convex optimization problems \cite{nesterov2017random}.

\begin{corollary}
\label{col:iid}
\textbf{(Global Convergence in \iid Setting)} 
Assuming the conditions of Theorem \ref{thm:iid} hold, the global convergence for \ours in the \iid case, characterized by $\Gamma = \frac{d - \zeta \gamma}{d \gamma}$, is given by
\begin{align}
\label{eq:global_satisfies_thm_iid}
\min_{t \in [T]} \mathbb{E}_t \Big\lVert \nabla f(\theta^{t}) \Big\rVert^2 &\leq \frac{f(\theta^{0}) - f^*}{2 \eta T \Gamma} + \frac{\sigma_g^2 \zeta L}{ NHd \Gamma} + \frac{\zeta \mu^2 L^3 }{4NH \Gamma}
\end{align}
where $f^*$ denotes the optimal loss value.
\end{corollary}

The upper bound on the minimum squared gradient norm across iterations is composed of three terms in Corollary \ref{col:iid}.
The first term indicates that the distance to the optimal loss relies on the initial state of optimality, while the second and third terms elucidate the influences of stochastic mini-batch errors and the perturbation scale $\mu$ inherent to ZOO, respectively. 
Specifically, they both reflect the impact of the model parameters $d$ and the low effective rank $r$ on the optimal loss. 
As pointed out in \cite{nesterov2017random}, by choosing an appropriate step size, we can obtain the desired accuracy. 

Given that $\gamma = \Theta(r)$ and $\zeta = \Theta\left(\frac{1}{rd}\right)$, we have $\zeta \gamma = \Theta\left(\frac{r}{rd}\right) = \Theta\left(\frac{1}{d}\right)$. As the parameter $d$ is large, $d-\zeta \gamma = d - \Theta\left(\frac{1}{d}\right)$ is dominated by $d$. Consequently, the dominant term of $\Gamma$ is $\frac{d \gamma}{d - \zeta \gamma}$, which simplifies to $\gamma = \Theta(r)$. Therefore, $\Gamma = \Theta\left(\frac{1}{\gamma}\right) = \Theta\left(\frac{1}{r}\right)$. Building on this relationship, we have the following corollary that articulates the convergence rate of \ours.

\begin{corollary}
\label{col:iid_rate}
\textbf{(Convergence Rate in \iid Setting)} 
Assuming the conditions of Corollary \ref{col:iid} hold and given $\eta = (NH)^{1/2} (rT)^{-1/2}$ and $\mu = (NH)^{1/4} r^{-1/2}$, we have
\begin{align}
\label{eq:global_rate_satisfies_thm_iid}
\min_{t \in [T]} \mathbb{E}_t \Big\lVert \nabla f(\theta^{t}) \Big\rVert^2 & \leq \mathcal{O}\left( r^{3/2} (NHT)^{-1/2} \right) + \mathcal{O}\left( d^{-1} (rNH)^{-1/2} \right).
\end{align}
\end{corollary}

The expression on the right-hand side of Eq. \eqref{eq:global_rate_satisfies_thm_iid} is dominated by $\mathcal{O}\left( r^{3/2} (NHT)^{-1/2} \right)$. Consequently, we have derived the convergence rate for FedMeZO. The low effective rank $r$ significantly contributes to lowering the convergence rate, which is also influenced by the number of clients $N$, the steps of local training iteration $H$, and the total number of communication rounds $T$. Moreover, to satisfy the learning rate condition in Eq. \eqref{eq:eta_satisfies_thm_iid}, the values of $N$, $H$, and $T$ must be suitably large.

It is important to note that \ours does not primarily aim to accelerate convergence speed but rather to identify the convergence rate under assumptions pertinent to LLMs. This is intended to demonstrate that \ours can achieve convergence even within a vast parameter space. In a series of studies on federated ZOO, Federated Zeroth-Order Optimization (FedZO) presents the most comprehensive and complete analysis with a convergence rate of $\mathcal{O}\left( \sqrt{d / (NHT b_1 b_2)} \right)$ \cite{fang2022communication}, which exhibits a lower rate compared to $\mathcal{O}\left(d^3 / T\right)$ of ZONE-S \cite{hajinezhad2019zone} and accounts for the impact of $H$ compared to $\mathcal{O}\left(\sqrt{d / NT}\right)$ of DZOPA \cite{yi2022zeroth}. In contrast to FedZO, our method, \ours, theoretically supports a faster convergence by replacing $d^{1/2}$ with $r^{3/2}$ and setting $b_1=b_2=1$.
These comparisons show that \ours addresses the challenges posed by large models, offering an efficient convergence rate that relies on $r$.

This advancement signifies progress in optimizing federated learning algorithms, particularly for LLMs, where the scalability of parameters and data heterogeneity are major challenges. By emphasizing the low effective rank, our approach enhances both the theoretical understanding of convergence behavior in complex settings and the guidance insights into the settings of learning rates and other parameters to achieve efficient convergence outcomes.

However, \iid data is typically encountered in idealized environments. In real-world applications, \noniid conditions are more common and challenging. Next, we further discuss and analyze the convergence of \ours under \noniid settings.

\subsection{Convergence Analysis in Non-i.i.d Case}
\label{subsec:result_noniid_convergence}
Analyzing convergence in the context of \noniid data distributions is crucial for understanding the behavior of FL algorithms in real-world scenarios. In this section, we extend our convergence analysis to the case where client data distributions are heterogeneous.

\begin{theorem}
\label{thm:noniid}
\textbf{(Stepwise Loss Descent in Non-i.i.d Setting)} 
Let Assumptions \ref{assumption:low_effective_rank}-\ref{assumption:minibatch} and Assumption \ref{assumption:local_global_noniid} hold and learning rate $\eta$ satisfy
\begin{equation}
\label{eq:eta_satisfies_thm_noniid}
\eta \leq \min \Bigl\{ \frac{1}{3 H L \sqrt{c_g d}}, \frac{N}{3 H L c_g}, \frac{1}{H^2} \Bigr\}.
\end{equation}
Then, the expected loss at each step for \ours in the non-i.i.d. setting is bounded as
\begin{align}
\label{eq:each_step_satisfied_thm_noniid}
\mathbb{E}_t \left[f(\theta^{t+1}) \right] &\leq f(\theta^{t}) - \left( \frac{2}{\gamma N} - \frac{2\zeta \widetilde{c}_h}{d} \right) \eta \mathbb{E}_t \left\lVert \nabla f(\theta^{t}) \right\rVert^2 \notag \\
&\quad + \frac{2 \widetilde{\sigma}^2 \zeta L \eta}{N H d} + \frac{\zeta \eta \mu^2 L^3 }{2 N H} - \frac{2}{\gamma N} \eta \sigma_h^2,
\end{align}
where $\widetilde{c}_h = c_h + N$ and $\widetilde{\sigma}^2 = 3 c_g \sigma_h^2 + \sigma_g^2$.
\end{theorem}

Comparing Eq. \eqref{eq:each_step_satisfied_thm_noniid} with its i.i.d. counterpart Eq. \eqref{eq:each_step_satisfies_thm_iid}, the non-i.i.d. setting introduces additional terms reflecting data heterogeneity. 
Firstly, an additional term $\widetilde{c}_h$ appears before $\mathbb{E}_t \lVert \nabla f(\theta^{t})\rVert^2$; secondly, the original $\sigma_g^2$ change into $\widetilde{\sigma}^2$; thirdly, a new term related to $\sigma_h^2$ is added at the end. 
The term $\widetilde{c}_h$ amplifies the effect of the gradient norm, while $\widetilde{\sigma}^2$ encapsulates both the intrinsic stochasticity and data heterogeneity. The presence of $\sigma_h^2$ indicates the impact of client data divergence on the convergence behavior, in a degree dependent on $\Theta(1/r d^2)$. 
Given that the contribution of the negative term accelerates the rate of decline in each round, it can be concluded that heterogeneity is positively correlated with convergence. Considering all the above changes, appropriate heterogeneity can aid in the model convergence.

Experimental results in Section \ref{subsubsec:exp_other} confirm that a more randomized dataset distribution leads to improved convergence, supporting our theoretical insights. Next, we present the global convergence result for the non-i.i.d. setting building upon Theorem \ref{thm:noniid}.

\begin{corollary}
\label{col:noniid}
\textbf{(Global Convergence in Non-i.i.d. Setting)} 
Assuming the conditions of Theorem \ref{thm:noniid} hold, denote $\widetilde{\Gamma} = \frac{d - N \gamma \zeta}{d \gamma N}$, \ours satisfies:
\begin{align}
\label{eq:global_satisfies_thm_noniid}
\min_{t \in [T]} \mathbb{E}_t \left\lVert \nabla f(\theta^{t}) \right\rVert^2 &\leq \frac{f(\theta^{0}) - f^*}{2 \widetilde{\Gamma} \widetilde{c}_h \eta T} + \frac{\widetilde{\sigma}^2 \zeta L}{\widetilde{\Gamma} \widetilde{c}_h N H d} \notag \\
&\quad + \frac{\zeta \mu^2 L^3 }{4 \widetilde{\Gamma} \widetilde{c}_h N H} - \frac{\sigma_h^2}{\widetilde{\Gamma} \widetilde{c}_h \gamma N}.
\end{align}
\end{corollary}

Given $\gamma = \Theta(r)$ and $\zeta = \Theta\left(\frac{1}{rd}\right)$, the expression $d - N \zeta \gamma$ simplifies to $d - \Theta\left(\frac{N}{d}\right)$ which is dominated by $d$. Consequently, $\widetilde{\Gamma}$ simplifies to $\Theta\left(\frac{1}{r}\right)$ as in the \noniid case.
Compared to Corollary \ref{col:iid}, Corollary \ref{col:noniid} introduces two changes: first, all terms on the right side of Eq. \ref{eq:global_satisfies_thm_noniid} include a denominator $c_g$, and second, there is an additional term associated with \noniid heterogeneity, scaled with $\Theta(\sigma_h^2 / c_h N)$. This further demonstrates the constraining effect of data heterogeneity in FedMeZO. 
Similar to Corollary \ref{col:iid_rate}, by setting appropriate values for $\eta$ and $\mu$, we obtain the following convergence rate.
\begin{corollary}
\label{col:noniid_rate}
\textbf{(Convergence Rate in Non-i.i.d. Setting)} 
Assuming the conditions of Corollary \ref{col:noniid} hold, with $\eta = (NH)^{1/2} (r \widetilde{c}_h T)^{-1/2}$ and $\mu = (\widetilde{c}h NH)^{1/4} r^{-1/2}$, \ours has convergence rate as follows:
\begin{align}
\label{eq:global_rate_satisfies_thm_noniid}
\min_{t \in [T]} \mathbb{E}_t \Big\lVert \nabla  & f(\theta^{t}) \Big\rVert^2 \leq \mathcal{O}\left( r^{3/2} (\widetilde{c}_h NHT)^{-1/2} \right) \notag \\
&\quad + \mathcal{O}\left( d^{-1} (r \widetilde{c}_h NH)^{-1/2} \right) - \mathcal{O}\left( \sigma_h^2 (c_h N)^{-1} \right).
\end{align}
\end{corollary}

The convergence rate in Eq. \eqref{eq:global_rate_satisfies_thm_noniid} is primarily driven by the term $\mathcal{O}\left( r^{3/2} (\widetilde{c}_h NHT)^{-1/2} \right)$, indicating that optimizing the balance between $\widetilde{c}_h$, $c_h$, and $N$ is crucial, which reflects a complex interplay of heterogeneity.
Specifically, we observe that to achieve better convergence, a smaller $\mathcal{O}\Bigl( r^{\frac{3}{2}} (\widetilde{c}_hNHT)^{-\frac{1}{2}} \Bigr)$ is preferred while the $\mathcal{O}\Bigl( \sigma_h^2 (c_h N)^{-1} \Bigr)$ term need to increase at the same time. Consequently, the balance between $c_g$, $c_h$, and $N$ becomes a dynamic trade-off process, \ie, the heterogeneity among different clients directly influences the overall convergence performance. 

For now, we have answered the question ``\textit{\textbf{Q2:} Can we establish the convergence properties of \ourtopic for LLMs?}'' via theorems and corollaries mentioned in this section. We also validated the nature of convergence under different scenarios and tasks through empirical experiments in Section \ref{subsec:exp_main_convergence}.

\subsection{Implications}
\label{subsec:result_implications}

The aforementioned theoretical results offer numerous insights into parameter tuning. A critical revelation from our analysis pertains to the constraints imposed on the learning rate, as delineated in Eq. \eqref{eq:eta_satisfies_thm_iid} and Eq. \eqref{eq:eta_satisfies_thm_noniid}, which suggests that an optimal learning rate magnitude is anchored at $1 /\sqrt{d}$. Larger learning rates are not only ineffectual but also pose a risk of destabilizing the training dynamic. In Appendix \ref{appendix:exp_lr_require}, our empirical experiments corroborate this hypothesis,  demonstrating that excessive learning rates precipitate abrupt increases in loss.

Furthermore, our insights regarding the learning rate open up prospects for personalized FL, a compelling approach that uses client-specific configurations to address heterogeneity and has attracted increasing interest \citep{sattler2020clustered, chen2022pflbench,fallah2020personalized, marfoq20neurips,chen2023efficient}. 
Specifically, we investigate theory-guided personalized strategies by dynamically adjusting the learning rate $\eta_i$ in proportion to a quantifiable measure of data heterogeneity among clients. 
In light of Theorem \ref{thm:noniid} that a larger heterogeneity is more conducive to model convergence, it is feasible to appropriately increase the learning rate, allowing specific clients to contribute more to the overall convergence, we propose the following tailored adjustment strategy:

\begin{proposition}
\label{prop:strategy}
\textnormal{(Adaptive Learning Rate Adjustment)} Let Assumption \ref{assumption:local_global_noniid} hold, the learning rate $\eta_i$ can be adjusted according to the formula to better accommodate the varied learning landscapes than non-personalized FL:
\begin{equation}
\label{eq:strategy}
    \eta_i = \eta_0(1+\alpha \cdot \Phi_i),
\end{equation}
\end{proposition}
\noindent where $\eta_0$ represents a default learning rate applicable in a \iid setting, $\alpha$ is a scaling factor that determines the sensitivity of the learning rate and $\Phi_i$ is the heterogeneity index, representing the extent of $c_g$ and $\sigma_h^2$. This proposition underscores the importance of considering data heterogeneity in the design of the learning rate strategy within personalized FL, offering a structured approach to enhance learning outcomes across diverse client datasets. 

In Section \ref{subsec:exp_personalized_lr}, we empirically confirm that a particular implementation of this strategy facilitates faster convergence.
It is important to note that the data heterogeneity index $\Phi_i$ cannot be determined a priori; therefore, we utilize several proxy measures during the training process to estimate it. 
Our goal is not to prescribe an exact solution to this strategy, but rather, through analysis and empirical investigation, to enlighten further research and development of personalized \ours for more effective training of LLMs.

These discussions and corresponding empirical support address the question ``\textbf{Q3}: \textit{Which model parameters are critical for convergence, and how can we leverage them to optimize FL performance, such as via personalization?}''.

Besides, recall that we adopt LoRA to mitigate the communication burden associated with LLMs for practical FL scenarios. Nonetheless, the influence of LoRA on the model's low effective rank, remains an open question. We thus advance the following conjecture under Assumption \ref{assumption:low_effective_rank}, predicated on existing literature\cite{yao2020pyhessian, sagun2017empirical}, to facilitate further validations:

\begin{conjecture}[Rank Correlation]
The optimal reparametrization rank $r_{\text{LoRA}}$ used in Low-Rank Adaptation (LoRA) is positively proportional to the effective rank $r$ of the Hessian matrix $\mathcal{H}(\theta^t)$ of the tuned LLM. The $r_{\text{LoRA}}$ is lower-bounded by $r$, and can serve as an empirical proxy for $r$.
\end{conjecture}

\section{Proof Outline}

This section provides an outline of the derivations presented in Section \ref{sec:main_results}, emphasizing the key analytical techniques and concepts employed. Detailed proofs are available in Appendix \ref{appendix:theorem}. 

We begin by taking expectations on both sides of Eq. \eqref{eq:pre_each_step_low_rank_estimator}, considering a federated learning setting. The equation is split into two main parts for further analysis:
\begin{align}
\small
    \label{eq:each_step_start_with}
    \mathbb{E}_t \bigl[ f(\theta^{t+1}) \bigr] \leq f(\theta^{t}) - \frac{1}{\gamma} \cdot \eta \mathbb{E}_t \Big\lVert \frac{1}{N} \sum_{i=1}^N \nabla f_i(\theta^{t}) \Big\rVert^2 \notag \\
    + \frac{1}{2} \eta^2 L \cdot \zeta \cdot \mathbb{E}_t \Big\lVert \frac{1}{N} \sum_{i=1}^N \sum_{k=1}^H e_i^{(t,k)} \Big\rVert^2.
\end{align}
For simplicity, we denote the two expectation terms as $T_1$ and $T_2$, which pertain to the expected squared norms of the gradient and the zeroth-order gradient estimator, respectively.

\subsection{Proof of Theorem \ref{thm:iid}}
\label{subsec:thm_iid}

For term $T_1$ in Eq. \eqref{eq:each_step_start_with}, we utilize the Cauchy-Schwarz inequality ${\lVert a + b \rVert}^2 \leq 2{\lVert a \rVert}^2 + 2{\lVert b \rVert}^2$ to decompose it into two parts, with the first representing the discrepancy between local and global gradients. By invoking Assumption \ref{assumption:local_global_iid}, we establish:
\begin{equation}
\label{eq:t1_iid}
T_1 \leq \frac{2}{N} \sum_{i=1}^N \mathbb{E}_t {\Big\lVert \nabla f_i(\theta^{t}) - \nabla f(\theta^{t}) \Big\rVert}^2 \notag + 2\cdot \mathbb{E}_t {\big\lVert \nabla f(\theta^{t}) \big\rVert}^2.
\end{equation}

For term $T_2$, Jensen's inequality allows us to bound the expected squared norm of the zeroth-order gradient estimator as follows:
\begin{equation}
\label{eq:t2_jensen_iid}
    T_2 \leq \frac{1}{N} \sum_{i=1}^N \sum_{k=1}^H \mathbb{E}_t \Big\lVert  e_i^{(t,k)} \Big\rVert^2.
\end{equation}

By substituting Eq.~\eqref{eq:pre_two_point_estimator_def} into Eq. \eqref{eq:t2_jensen_iid}, we proceed to use the Cauchy-Schwarz inequality to decompose this gradient estimator into two parts, each of which is a biased estimator. 
Recall that in our gradient estimator, $z_i$ follows a Gaussian distribution. Therefore, the impact on the norm caused by a forward step and a backward step of the estimator is identical. 
Consequently, we ascertain that the term $T_2$ is bounded by a single-point gradient estimation as $\mathbb{E}_t \Bigg\lVert  \frac{z_i^{(t,k)}}{\mu}\Bigl( F_i(\theta_i^{(t,k)}+\mu z_i^{(t,k)}, \mathcal{B}_i^{(t,k)})$ $ -F_i(\theta_i^{(t,k)}, \mathcal{B}_i^{(t,k)}) \Bigr) \Bigg\rVert^2$.

Following Lemma 4.1 in \cite{gao2018information}, we can bound the expectation term as:
\begin{align}
\label{eq:t2_trans_iid}
    \mathbb{E}_t \Big\lVert  e_i^{(t,k)} & \Big\rVert^2 \leq \frac{1}{d^2} \Biggl[ 2d\cdot\mathbb{E}_t \Big\lVert \nabla F_i(\theta_i^{(t,k)}, \mathcal{B}_i^{(t,k)})\Big\rVert^2 + \frac{\mu^2}{2} L^2 d^2 \Biggr]\notag \\
    &\leq \frac{1}{d^2} \Biggl[ 2 c_g d\cdot \mathbb{E}_t \Big\lVert \nabla f_i(\theta_i^{(t,k)})\Big\rVert^2 + 2 d \sigma_g^2+ \frac{\mu^2}{2} L^2 d^2 \Biggr],
\end{align}
where the second inequality is derived based on Assumption \ref{assumption:minibatch}.
Subsequently, we bound the expectation term by applying the Cauchy-Schwartz inequality to divide it into three parts: 
\begin{align}
\label{eq:t2_fi_iid}
    \mathbb{E}_t \Big\lVert \nabla f_i(\theta_i^{(t,k)})\Big\rVert^2 &= \mathbb{E}_t \Big\lVert \nabla f_i(\theta_i^{(t,k)}) \mp \nabla f_i(\theta^t) \mp \nabla f(\theta_i^t) \Big\rVert^2 \notag \\
    & \leq 3 L^2 \mathbb{E}_t \Big\lVert \theta_i^{(t,k)} - \theta^t \Big\rVert^2 + 3\mathbb{E}_t \Big\lVert \nabla f(\theta^t) \Big\rVert^2.
\end{align}
The first part represents the gradient difference between stages $(t, k)$ and $(t, 0)$, which can be computed using Assumption \ref{assumption:lsmooth}, \ie, the $L$-smooth condition. The second part signifies the disparity between local and global aspects, calculated using Assumption \ref{assumption:local_global_iid}. The third part is retained as is.

Combining Equations \eqref{eq:t2_jensen_iid}, \eqref{eq:t2_trans_iid} and \eqref{eq:t2_fi_iid}, we bound $T_2$ as follow:
\begin{align}
\label{eq:t2_final_iid}
    T_2 \leq \frac{6 c_g L^2}{Nd} &\sum_{i=1}^N \sum_{k=1}^H \mathbb{E}_t \Big\lVert \theta_i^{(t,k)} - \theta^t \Big\rVert^2  \notag \\
    + &\frac{6 c_gH}{d} \mathbb{E}_t \Big\lVert \nabla f(\theta^t) \Big\rVert^2  + \frac{2 H \sigma_g^2}{d} + \frac{\mu^2 H L^2}{2}.
\end{align}

Next, combining Equations \eqref{eq:each_step_start_with}, \eqref{eq:t1_iid} and \eqref{eq:t2_final_iid}, we have:
\begin{align}
\label{eq:each_step_t1t2_iid}
    \mathbb{E}_t \bigl[ f(\theta^{t+1}) \bigr]  - f(\theta^{t}) \notag 
    &\leq \Biggl( \frac{3 c_g \zeta \eta^2 H L}{d} - \frac{2 \eta}{\gamma}  \Biggr) \mathbb{E}_t \Big\lVert \nabla f(\theta^{t}) \Big\rVert^2 \notag \\
    &+ \frac{3 c_g \zeta \eta^2 L^3}{N d} \sum_{i=1}^N \sum_{k=1}^H \mathbb{E}_t \Big\lVert \theta_i^{(t,k)} - \theta^t \Big\rVert^2 \notag \\
    &+ \frac{\sigma_g^2 \zeta \eta^2 H L}{d} + \frac{\zeta \eta^2 \mu^2 H L^3 }{4}.
\end{align}

In Eq. \eqref{eq:each_step_t1t2_iid}, $\mathbb{E}_t \Big\lVert \theta_i^{(t,k)} - \theta^t \Big\rVert^2$ remains unknown and we need to constrain it further. 
The key idea is to transform this expectation term into a form related to $\mathbb{E}_t^k \lVert e_i^{(t,k)} \rVert^2$ and then utilize the conclusion of Eq. \eqref{eq:t2_trans_iid} and Eq. \eqref{eq:t2_fi_iid} for computation. The detailed derivation process is provided in the Appendix \ref{appendix:thm_iid} and we can have the bounded result:
\begin{align}
\label{eq:lemma2_final_iid}
    &\frac{1}{N} \sum_{i=1}^N \sum_{k=1}^H \mathbb{E}_t \Big\lVert \theta_i^{(t,k)} - \theta^t \Big\rVert^2 \leq \frac{C_1}{C_0}, 
\end{align}
where $C_0 = 1-3 c_g d \eta^2 H^2 L^2$ and $C_1=2 c_g d H^3 \eta^2 \mathbb{E}_t \big\lVert \nabla f(\theta^t) \big\rVert^2$ $+ \frac{2}{3}d\sigma_g^2 H^3 \eta^2 + \frac{\mu^2 L^2 d^2 H^3 \eta^2}{6}$.

Finally, by substituting Eq. \eqref{eq:lemma2_final_iid} into Eq. \eqref{eq:each_step_t1t2_iid}, we obtain the final result of the stepwise descent. After simplifying and appropriately setting the learning rate, we arrive at the result presented in Theorem \ref{thm:iid}. The detailed derivation of this result is provided in Appendix \ref{appendix:thm_iid}.

\subsection{Proof of Theorem \ref{thm:noniid}}
The proof for the \noniid case follows a similar structure to that of the \iid case, with adjustments made for the heterogeneity between local and global models as captured by $c_h$ and $\sigma_h^2$ (Assumption \ref{assumption:local_global_noniid}). 
In particular, we redefine $T_1$ in Eq. \eqref{eq:each_step_start_with} as $\widetilde{T}_1$ to reflect the increased variance due to \noniid data:
\begin{align}
    \label{eq:t1_noniid}
    \widetilde{T}_1 &\leq \frac{2}{N} \sum_{i=1}^N \mathbb{E}_t {\Big\lVert \nabla f_i(\theta^{t}) - \nabla f(\theta^{t}) \Big\rVert}^2 + 2\mathbb{E}_t {\big\lVert \nabla f(\theta^{t}) \big\rVert}^2 \notag \\
    &\leq 2(1+c_h) \mathbb{E}_t {\big\lVert \nabla f(\theta^{t}) \big\rVert}^2 + 2 \sigma_h^2,
\end{align}
where the second inequality follows Assumption \ref{assumption:local_global_noniid}.

Subsequently, $\widetilde{T}_2$ is computed similarly, with the heterogeneity terms incorporated. The difference from the \iid case lies in the bounding of expectation term in Eq. \eqref{eq:t2_fi_iid}:
\begin{align}
\label{eq:t2_fi_noniid}
    \mathbb{E}_t \Big\lVert \nabla f_i(\theta_i^{(t,k)})\Big\rVert^2 &\leq 3 L^2 \mathbb{E}_t \Big\lVert \theta_i^{(t,k)} - \theta^t \Big\rVert^2 \notag \\
    &\qquad + 3(c_h+1)\mathbb{E}_t \Big\lVert \nabla f(\theta^t) \Big\rVert^2 + 3\sigma_h^2,
\end{align}
where the inequality employs Assumption \ref{assumption:local_global_noniid}.

Combining Equations \eqref{eq:t2_jensen_iid}, \eqref{eq:t2_trans_iid} and \eqref{eq:t2_fi_noniid}, we bound $\widetilde{T}_2$ as follow:
\begin{align}
\label{eq:t2_final_noniid}
    \widetilde{T}_2 \leq \frac{6 c_g L^2}{Nd} \sum_{i=1}^N \sum_{k=1}^H & \mathbb{E}_t \Big\lVert \theta_i^{(t,k)} - \theta^t \Big\rVert^2  
    + \frac{6 c_g (c_h+1)H}{d} \mathbb{E}_t \Big\lVert \nabla f(\theta^t) \Big\rVert^2 \notag \\
    & + \frac{6 c_g \sigma_h^2 H}{d}+ \frac{2 H \sigma_g^2}{d} + \frac{\mu^2 H L^2}{2}.
\end{align}

By substituting $\widetilde{T}_1$ in Eq. \eqref{eq:t1_noniid} and $\widetilde{T}_2$ in Eq. \eqref{eq:t2_final_noniid} into Eq. \eqref{eq:each_step_start_with}, we get the result under the \noniid condition as follow: 
\begin{align}
\label{eq:each_step_t1t2_noniid}
    \mathbb{E}_t \bigl[ f(\theta^{t+1}) \bigr] - f(\theta^{t}) 
    &\leq \Biggl( \frac{3 c_g \widetilde{c}_h \zeta \eta^2 H L}{d} - \frac{2 \widetilde{c}_h \eta}{\gamma } \Biggr) \mathbb{E}_t \Big\lVert \nabla f(\theta^{t}) \Big\rVert^2 \notag \\
    &+ \frac{3 c_g \zeta \eta^2 L^3 }{N d} \sum_{i=1}^N \sum_{k=1}^H \mathbb{E}_t \Big\lVert \theta_i^{(t,k)} - \theta^t \Big\rVert^2 \notag \\
    &+ \frac{\widetilde{\sigma}^2 \zeta \eta^2 H L}{d} + \frac{\zeta \eta^2 \mu^2 H L^3 }{4} - \frac{2}{\gamma} \sigma_h^2 \eta,
\end{align}
where $\widetilde{c}_h$ denotes $(c_h+1)$ and $\widetilde{\sigma}^2$ denotes $(3c_g \sigma_h^2 + \sigma_g^2)$.
We then need to make the expectation term bounded. Unlike Eq. \eqref{eq:lemma2_final_iid}, due to the variations introduced by Eq. \eqref{eq:t2_fi_noniid}, two additional terms related to $\widetilde{c}_h$ and $\widetilde{\sigma}^2$ emerge, yielding the following result:
\begin{align}
\label{eq:lemma2_final_noniid}
    \frac{1}{N} \sum_{i=1}^N \sum_{k=1}^H \mathbb{E}_t \Big\lVert \theta_i^{(t,k)} - &\theta^t \Big\rVert^2 \leq \frac{C_2}{C_0},
\end{align}
where $C_1= 1-3 c_g d \eta^2 H^2 L^2$ and $C_2=2 c_g d \widetilde{c}_h H^3 \eta^2 \mathbb{E}_t \big\lVert \nabla f(\theta^t) \big\rVert^2$ $+ \frac{2}{3} \widetilde{\sigma}^2 d H^3 \eta^2 + \frac{\mu^2 L^2 d^2 H^3 \eta^2}{6}$.

Finally, we can substitute Eq. \eqref{eq:lemma2_final_noniid} into Eq. \eqref{eq:each_step_t1t2_noniid} and have the result of Theorem \ref{thm:noniid}.
The detailed derivations about these steps of Theorem \ref{thm:noniid} are provided in Appendix \ref{appendix:thm_noniid}.

\section{Empirical Support}
\label{sec:exp}

This section aims to empirically validate our theoretical findings through a series of experiments.

\subsection{Experimental Setup}
\label{subsec:exp_setup}
We utilize LLaMA-3B \cite{touvron2023llama} as the foundational model and employ four datasets covering a range of tasks and data distribution types to provide comprehensive validation of our theoretical results \cite{kuang2023federatedscope-LLM}. Given that our theory centers on the loss function, we primarily focus on analyzing loss descent in our experiments. More details of the used datasets are in the Appendix (Table \ref{table:datasets}).

We set the total number of communication rounds to 500. By default, BP-based baselines undertake local training for one epoch, whereas \ours conducts local training for 30 steps. 
We repeat our experiments with three seeds and plot the error bars. 
For more detailed implementation specifics, please refer to Appendix \ref{appendix:exp_setup}.

\subsection{Convergence Study}
\label{subsec:exp_main_convergence}
To assess the convergence of \ours, we perform experiments on three datasets using different data splitters, as specified in Table \ref{table:datasets}, with test loss serving as the convergence metric. 
Our objective is to evaluate the generalization and stability of \ours across diverse datasets and heterogeneity scenarios. For benchmarking purposes, we also measure the performance of BP-based FedAvg on the same datasets. 
Additionally, we document the GPU memory usage during training in Table \ref{tab:gpu_memory}. 
Representative findings are illustrated in Figure \ref{fig:main_convergence}, while all the comprehensive results are available in Appendix \ref{appendix:exp_main_convergence_app} due to the space limitation.

\begin{figure}[h!]
    \centering
    \includegraphics[width=\columnwidth]{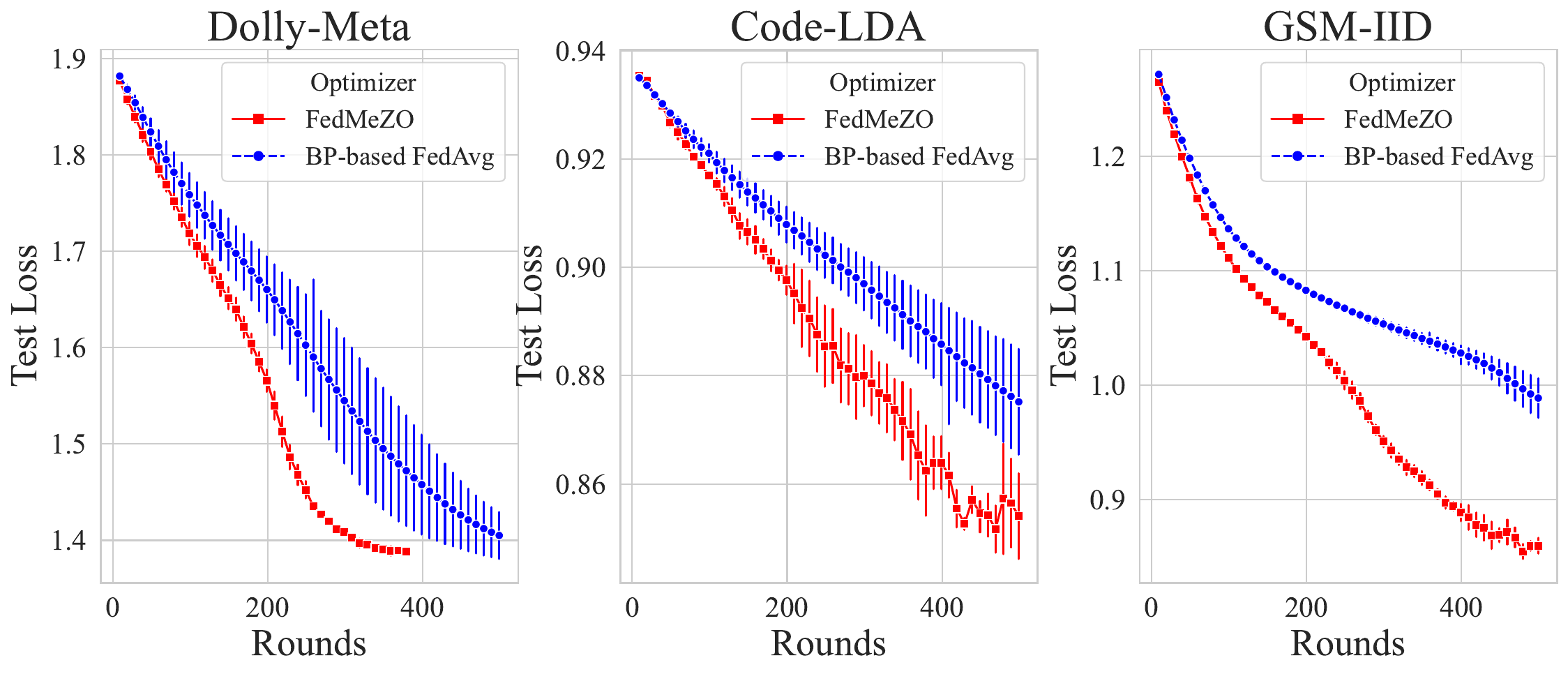}
    \caption{Convergence comparison of \ours and BP-based FedAvg. More results are in Appendix \ref{appendix:exp_main_convergence_app}.}
    \label{fig:main_convergence}
\end{figure}

\begin{table}[h]
\caption{The GPU Memory of BP-based FedAvg and \ours.}
\label{tab:gpu_memory}
\begin{tabular}{@{}ccc@{}}
\toprule
Task & BP-based FedAvg (MiB) & \ours (MiB) \\ \midrule
Dolly-Meta & 26571   & 10061  \\
GSM8K-IID & 17771 & 9733 \\
CodeAlpaca-LDA & 15287 & 9569 \\ \bottomrule
\end{tabular}
\end{table}
Two main conclusions emerge from the convergence experiments: First, when the learning rate complies with the requirements discussed in Section \ref{subsec:result_implications}, as stipulated by Theorem \ref{thm:iid} and Theorem \ref{thm:noniid}, \ours consistently diminishes loss with each step, ultimately achieving stable convergence. Second, under equivalent learning rate configurations, \ours decreases loss more rapidly than BP-based FedAvg, indicating a swifter convergence rate. For instance, in the Dolly-Meta figure, \ours stabilizes and converges around 300 rounds, whereas BP-based FedAvg's loss is still declining at this juncture. Notably, from Table \ref{tab:gpu_memory}, we observe that the GPU memory demand for \ours is roughly one-half of that required by BP-based FedAvg, suggesting that \ours can achieve a speedier convergence with fewer resources.

\subsection{Hyper-parameters Study}
\label{subsec:hyperpara}
In this subsection, we perform a series of experiments to ascertain the influence of various hyper-parameters, as intimated by our theoretical findings.

\subsubsection{Impact of Perturbation Scale $\mu$}
\label{subsubsec:exp_impact_mu}

To corroborate the theoretical impacts of the perturbation step on convergence, we examine $\mu$ values of $5 \times 10^{-3}$ and $2 \times 10^{-4}$, in addition to the default $\mu = 1 \times 10^{-3}$. We leverage the same datasets and splitters as in Section \ref{subsec:exp_main_convergence} for robustness. Figure \ref{fig:diff_mu} shows representative outcomes, with comprehensive results in Appendix \ref{appendix:exp_impact_mu}.

\begin{figure}[h]
\centering
\includegraphics[width=\columnwidth]{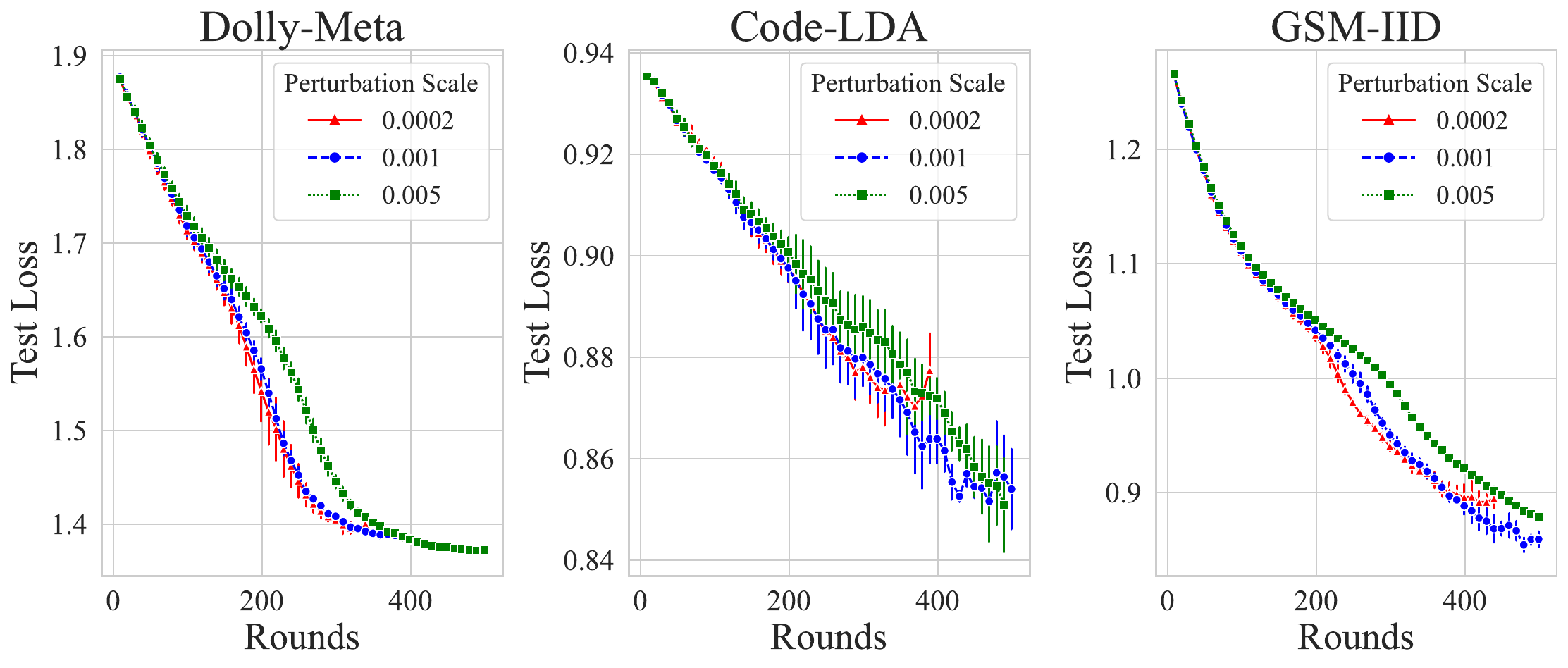}

\caption{Effects of different perturbation scales $\mu$. More results are in Appendix \ref{appendix:exp_impact_mu}.}
\label{fig:diff_mu}
\end{figure}

The results confirm that, consistent with Equations \eqref{eq:global_rate_satisfies_thm_iid} and \eqref{eq:global_rate_satisfies_thm_noniid}, a smaller $\mu$ marginally expedites model convergence. Figure \ref{fig:diff_mu} exemplifies that the training trajectory with $\mu = 2 \times 10^{-4}$ descends more rapidly than the others. However, given that $\mu$ appears as a second-order term in $\frac{\zeta \mu^2 L^3 }{4 \widetilde{\Gamma} \widetilde{c}_h N H}$ and its absolute value is relatively small, its overall influence is modest. This is evident in Figure \ref{fig:diff_mu}, where modifications to $\mu$ within a specific range yield only slight variations. Thus, a smaller $\mu$ proves advantageous for model convergence.

\subsubsection{Impact of Local Iteration Step $H$}
\label{subsubsec:exp_impact_h}

To validate the theoretical impact of local iteration steps on convergence, we contrast $H=10$ and $H=50$ with the standard $H=30$. Utilizing identical datasets and splitters from Section \ref{subsec:exp_main_convergence}, we present typical findings in Figure \ref{fig:diff_H}, with all the detailed results forthcoming in Appendix \ref{appendix:exp_impact_H}.

\begin{figure}[h]
    \centering
    \includegraphics[width=\columnwidth]{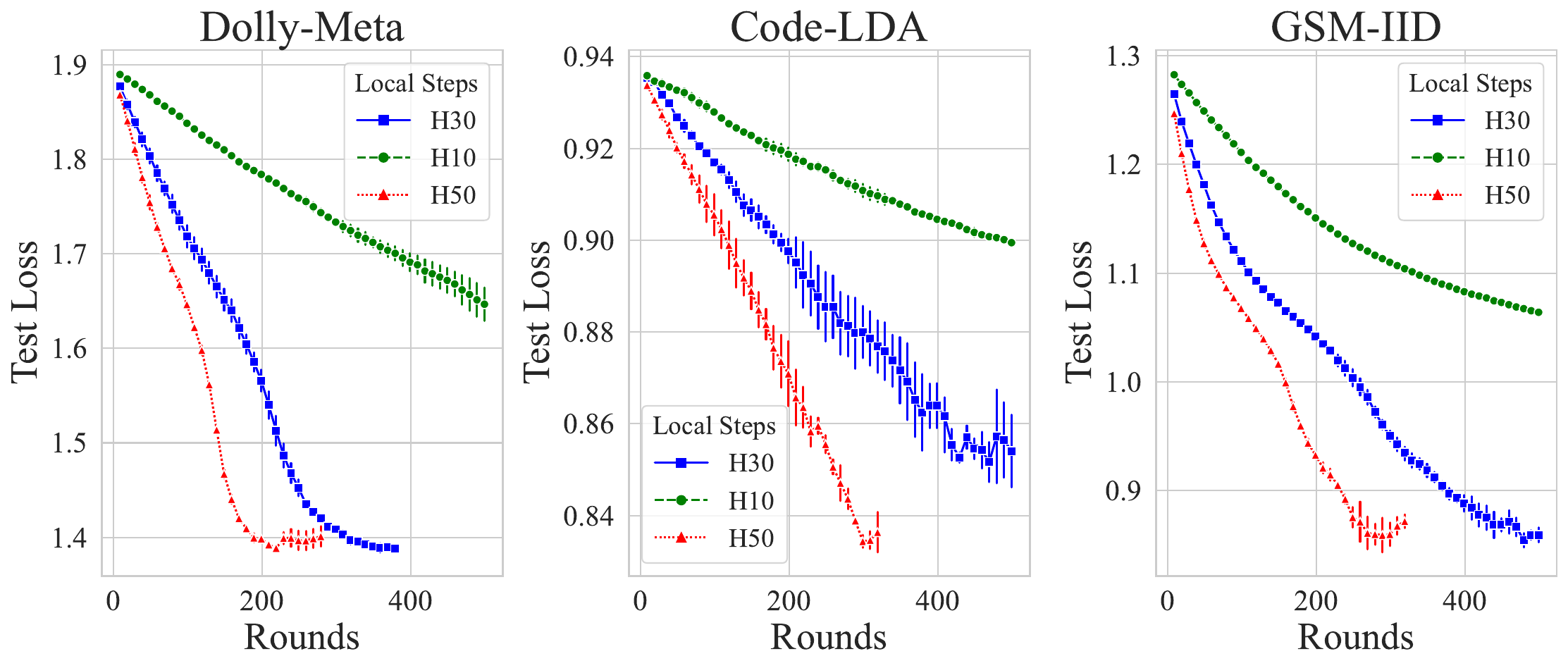}
\caption{Effects of different local iteration steps $H$. More results are in Appendix \ref{appendix:exp_impact_H}.}
    \label{fig:diff_H}
\end{figure}

These experimental results suggest that a lower $H$ engenders a more sluggish convergence pace, whereas a higher $H$ somewhat propels convergence, mirroring the impact of $H$ as a denominator in the theoretical convergence rate analysis. Nonetheless, an excessive $H$ may lead to instability, as depicted by the curve of $H=50$, which exhibits a surge endwise in the figure. Hence, an appropriate choice of $H$ facilitates efficient model convergence.

\subsubsection{Analysis of Other Hyper-parameters}
\label{subsubsec:exp_other}
We also explore the ramifications of learning rate, data splitters, the number of clients $N$ on the convergence rate, the batch size and the model size. Due to the space limit, details pertaining to these experimental settings and results are presented in Appendices \ref{appendix:exp_lr_require}, \ref{appendix:exp_impact_hetero}, \ref{appendix:exp_impact_client_num}, \ref{appendix:exp_impact_batch_size} and \ref{appendix:exp_impact_model_size} respectively. 

In a nutshell, as shown in Fig. \ref{fig:lr_require}, a suitable learning rate anchored at $1 /\sqrt{d}$ leads to stabilized training dynamics, while larger ones exhibit divergence as suggested by our analysis (Section \ref{subsec:result_implications}).
Besides, dissimilar splitters symbolize varying extents of data heterogeneity, and we have observations revealing that augmented heterogeneity culminates in lower stabilized loss values (as shown in Fig. \ref{fig:diff_split}). This intimates that a moderate degree of data heterogeneity can elevate the model's convergence proficiency.
Moreover, Fig. \ref{fig:diff_n} and Table \ref{table:client_num} verifies our theoretical analysis in Section \ref{sec:main_results} that an increase in $N$ helps stabilize the global convergence.

\subsection{Personalization Study}
\label{subsec:exp_personalized_lr}

To reconcile Proposition \ref{prop:strategy} with practical scenarios, we conduct the following subsequent experiments. To account for each client's heterogeneity during model updates in each round, we derive three signal quantities: (1) \textit{Round-wise Train Loss Difference}: The discrepancy between each client's loss in the preceding training round and the global loss. (2) \textit{Five-round Average Train Loss Difference}: The average loss deviation for each client relative to the global loss over the antecedent five rounds. (3) \textit{Model Parameter Update Difference}: The disparity between each client's previous round parameter updates and the global update magnitude. We normalize them to the range of $(-1, 1)$, serving as empirical estimates for $\Phi$.

For the setting of the scaling factor $\alpha$, following the guidance of the learning rate in Section \ref{subsec:result_implications}, we designate $1.5\times10^{-5}$ as the maximal learning rate, potentially leading to surges as per the learning rate search network. Symmetrically, we posit the minimal value at $5\times10^{-6}$, anchored on the default learning rate of $1\times10^{-5}$, thereby assigning $\alpha$ a value of $5\times10^{-6}$.

\begin{figure}[h]
    \centering
    \includegraphics[width=\columnwidth]{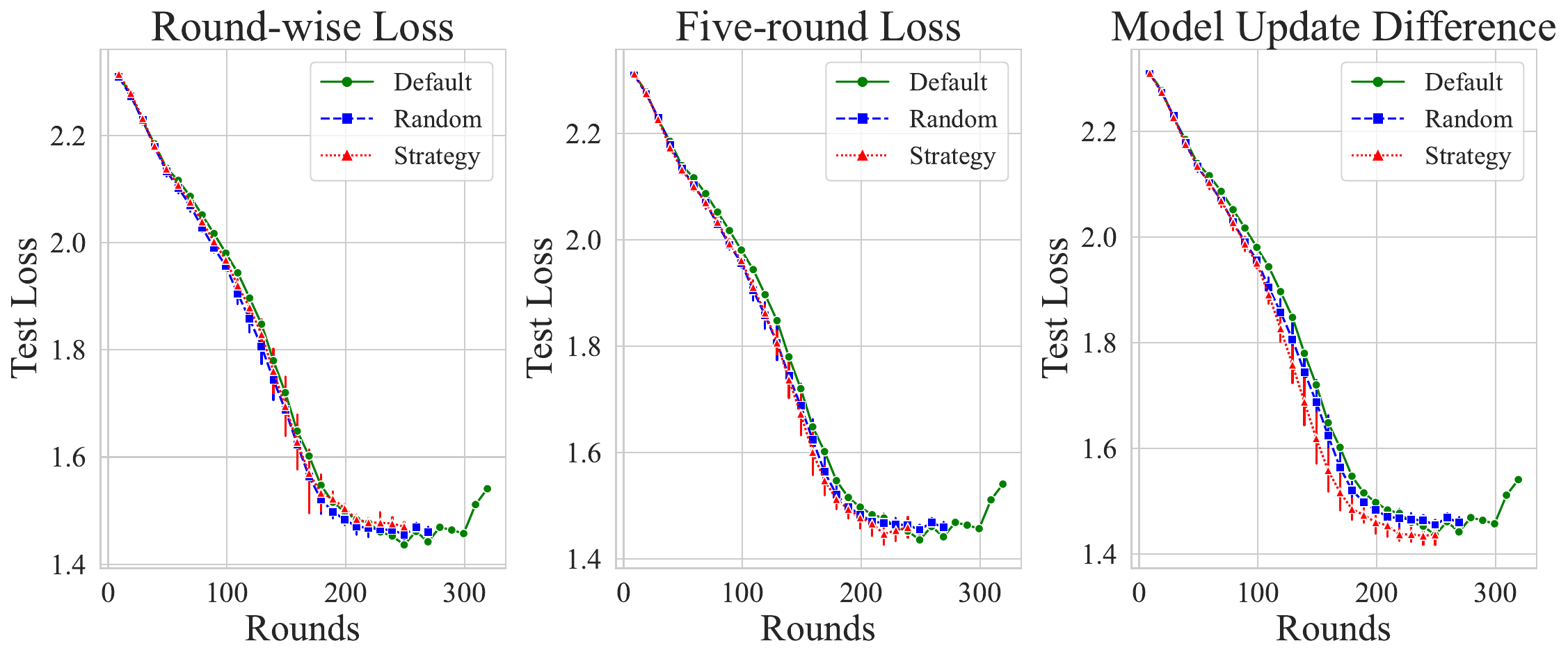}
            \vspace{-0.15in}
\caption{Comparison of different strategies of learning rate adjustment. ``Default'' indicates non-personalized case, and ``Round-wise Loss'', ``Five-round Loss'' and ``Model Update Difference'' indicate three signal quantities leveraged. }
    \label{fig:personalized_lr}
\end{figure}

As counterpoints, we furnish two configuration strategies for the learning rate adjustment: one uniformly applies a default learning rate of $1\times10^{-5}$, while the other randomly selects within $[5\times10^{-6}, 1.5\times10^{-5}]$ for each round. The conclusive results are depicted in Figure \ref{fig:personalized_lr}.
Based on the experimental results, we observe that our method achieves faster loss convergence with the second and third types of signal quantities compared to the default and random settings, with the third type yielding the most impressive performance. 
In contrast, the first type of signal quantity had a negligible impact. 
These results suggest that while round-wise loss exhibits some degree of randomness, aggregating losses over multiple rounds can approximate heterogeneity to a meaningful extent, thus serving as an indicator to expedite model convergence. 

It is also noteworthy that the third type of signal quantity aligns with the expression $\lVert \nabla f(\theta^t) - \sum_{k=0}^H \eta_i \nabla f_i(\theta^{(t,k)}) \rVert^2$, which most closely reflects Assumption \ref{assumption:local_global_noniid}. 
Consequently, it demonstrates the most effective performance in the experiments, not only achieving the fastest convergence but also the lowest stable loss. 
This case study experiment substantiates the efficacy of Proposition \ref{prop:strategy}, offering valuable insights for parameter tuning in personalized FL.

\section{Conclusion}
\label{sec:conclusion}

This work investigates the convergence of FedMeZO, a practical approach integrating Memory-efficient Zeroth-Order Optimization within a federated learning setting for Large Language Models (LLMs). Extensive empirical results verified our analyses and indicated that FedMeZO achieves fast convergence with reduced GPU memory requirements, offering a promising alternative to traditional optimization methods. The incorporation of a personalized learning rate adjustment, derived from theoretical analysis, has been shown to effectively enhance loss reduction. 
Through this study, we aim to enlighten more research and development of memory-efficient optimization techniques to address practical challenges associated with the fine-tuning of LLMs, particularly in resource-constrained scenarios \cite{hu2022lora}.






\bibliographystyle{ACM-Reference-Format}
\bibliography{FedMeZO}

\newpage
\appendix

\section*{Appendix}
Our appendix is organized as follows:
\begin{itemize}
    \item Section \ref{appendix:notation} summarizes all the used mathematical symbols.
    \item Section \ref{appendix:assumptions} gives the introduced FL-related assumptions for our analysis.
    \item Section \ref{appendix:lemma_proof} and Section \ref{appendix:theorem} describe the detailed proof procedures of our lemmas and theorems respectively.
    \item Section \ref{appendix:exp_setup} details our empirical implementations in terms of the datasets, platforms, and hyper-parameters.
    \item Section \ref{appendix:exp_all} presents additional experiment results about the evaluations with common LLM's metrics (Section \ref{appendix:exp_llm_metrics}), the conditions of learning rate (Section \ref{appendix:exp_lr_require}), the impact of data heterogeneity (Section \ref{appendix:exp_impact_hetero}), the impact of client number (Section \ref{appendix:exp_impact_client_num}), the impact of batch size (Section \ref{appendix:exp_impact_batch_size}), the impact of model size (Section \ref{appendix:exp_impact_model_size}), the impact of perturbation scale (Section \ref{appendix:exp_impact_mu}), the impact of local iterations (Section \ref{appendix:exp_impact_H}), and the full results of convergence curves (Section \ref{appendix:exp_main_convergence_app}).

\end{itemize}

\section{Notation}
\label{appendix:notation}

For ease of reading and reference, we present all mathematical symbols used in this paper in Table \ref{table:symbols}.

\begin{table*}[htbp]
\centering
\caption{Description of symbols used in the paper.}
\label{table:symbols}
\begin{tabular}{|c|l|}
\hline
\textbf{Symbol}                                     & \textbf{Description}                               \\ \hline
$f(\theta)$                                         & Global loss function over the parameter $\theta$                                  \\ \hline
$\nabla f(\theta)$                                  & Gradient of the loss function with respect to parameter $\theta$                             \\ \hline
$F_i(\theta, \mathcal{B}_i)$ & Local loss function on the $i^{th}$ client with mini-batch $\mathcal{B}_i$ \\ \hline
$\nabla F(\theta, \mathcal{B})$                     & The gradient of parameter $\theta$ with mini-batches           \\ \hline
$\widetilde{\nabla} F_i(\theta, z, \mathcal{B}_i, \mu)$ & Zeroth-order gradient estimator for $F_i$ with perturbation $\mu$ \\ \hline
$z$ & Gaussian random variable sampled from $\mathcal{N}(0, I_d)$ \\ \hline
$\mathcal{B}_i$ & Mini-batch of data sampled from the local distribution $\mathcal{D}_i$ \\ \hline
$\mu$ & Perturbation scale for zeroth-order gradient estimation \\ \hline
$\eta$ & Learning rate for model updates \\ \hline
$n$ & Number of perturbations in $n$-SPSA zeroth-order optimization \\ \hline
$(t,k)$ & $k^{th}$ iteration within the $t^{th}$ communication round \\ \hline
$\mathcal{H}$ & Hessian matrix of the loss function \\ \hline
$r$ & The low effective rank of the Hessian matrix \\ \hline
$\gamma$ & Factor quantifying the effective low-rank property of the gradient \\ \hline
$\zeta$ & Factor quantifying the effective low-rank property of the gradient estimator \\ \hline
$d$ & Dimension of the model parameter $\theta$ \\ \hline
$L$ & Smoothness constant of the loss function \\ \hline
$N$ & Number of clients participating in FL \\ \hline
$i$ & Index identifying the $i^{th}$ client \\ \hline
$T$ & Number of communication rounds in FL \\ \hline
$e_i^{(t,k)}$ & Zeroth-order gradient estimator of the $i^{th}$ client in iteration $(t,k)$ \\ \hline
$H$ & Total number of local iterations within a communication round \\ \hline
$c_g$, $\sigma_g$ & Constants related to the gradient estimation gap caused by mini-batch stochasticity \\ \hline
$c_h$, $\sigma_h$ & Constants related to the heterogeneity of client data and the global model \\ \hline
$\mathbb{E}_t$ & Expectation taken over the randomness in the $t^{th}$ round \\ \hline
$\mathbb{E}_t^k$ & Expectation taken over the randomness in the $k^{th}$ iteration of the $t^{th}$ round \\ \hline
\end{tabular}
\end{table*}

\section{Detailed Assumptions}
\label{appendix:assumptions}

\begin{assumption}
\label{assumption:lower_bound}
    \textbf{(Bounded Loss)} 
    The global loss function $f(\theta)$ is bounded below by a scalar $f^*$, i.e., $ f^* \geq f(\theta) > -\infty$ for all $\theta$.
\end{assumption}

\begin{assumption}
\label{assumption:lsmooth}
    \textbf{($L$-smoothness)} 
    The local and global loss functions $F_i(\theta, \mathcal{B}_i)$, $f_i(\theta)$, and $f(\theta)$ are L-smooth. Mathematically, for any $\theta_1, \theta_2 \in \mathbb{R}^d$, it holds that
    
    \begin{align}
        \lVert \nabla f_i(\theta_2) - \nabla f_i(\theta_1) \rVert &\leq L \lVert \theta_2 - \theta_1 \rVert, \notag \\
        f_i(\theta_2) \leq f_i(\theta_1) + \langle \nabla f_i(\theta), \theta_2 &- \theta_1 \rangle + \frac{L}{2} {\lVert \theta_2 - \theta_1 \rVert}^2. \notag
    \end{align}
    
\end{assumption}

\begin{assumption}
\label{assumption:minibatch}
     \textbf{(Mini-batch Gradient Error Bound)} 
     For any $\theta \in \mathbb{R}^d$, the second-order moment of the stochastic gradient is bounded by $\mathbb{E}_{\mathcal{B}_i} \lVert \nabla F_i(\theta, \mathcal{B}_i) \rVert^2 \leq c_g \lVert \nabla f_i(\theta) \rVert^2 + \sigma_g^2$, where $c_g \geq 1$.
\end{assumption}

\begin{assumption}
\label{assumption:local_global_iid}
     \textbf{(Global-Local Disparities in \iid Setting)} 
     For any $\theta \in \mathbb{R}^d$, the discrepancy between the local and global gradient is negligible, i.e., $\mathbb{E}_i \lVert \nabla f(\theta) - \nabla f_i(\theta) \rVert^2 = 0$.
\end{assumption}

\begin{assumption}
\label{assumption:local_global_noniid}
    \textbf{(Global-Local Disparities in Non-i.i.d. Setting)} 
    For any $\theta \in \mathbb{R}^d$, the discrepancy between the local and global gradient is bounded by $\lVert \nabla f(\theta) - \nabla f_i(\theta) \rVert^2 \leq c_h \lVert \nabla f(\theta) \rVert^2 + \sigma_h^2,$ where $c_h$ is a positive constant.
\end{assumption}

Assumptions \ref{assumption:lower_bound}-\ref{assumption:minibatch} are well-established in the literature on large-scale stochastic optimization \cite{bottou2018optimization}. Assumption \ref{assumption:local_global_iid} describes an ideal \iid setting where each client's gradient is aligned with the global gradient. Assumption \ref{assumption:local_global_noniid} accounts for the heterogeneity of client data distributions that is typical in \noniid settings \citep{li2019convergence, wang2021novel, li2020federated}. 

\section{Full Proof of Lemma}
\label{appendix:lemma_proof}

\subsection{Proof of Lemma \ref{lemma:unbiased}}
\label{appendix:lemma_unbiased}
We recap the proof of Stein's identity following \cite{feng2023does}, noting that $\delta$ is a random variable sampled from $\mathcal{N}(0, \sigma^2 I_d)$, and we replace it with $\mu z_i$ in Definition \ref{def:two_point_zoo}.

\begin{align}
    \nabla F_i(\theta, \mathcal{B}_i) &= \nabla \mathbb{E}_{\delta \sim \mathcal{N}(0, \sigma^2 I_d)} \Bigl[ F_i(\theta + \delta, \mathcal{B}_i) \Bigr] \notag \\
    &= (2 \pi)^{-\frac{d}{2}} \nabla \int F_i(\theta + \delta, \mathcal{B}_i) \cdot \textit{exp} \Bigl( - \frac{\lVert \delta \rVert^2}{2 \sigma^2} \Bigr) d \delta \notag \\
    &= (2 \pi)^{-\frac{d}{2}} \int F_i(\widetilde{\theta}, \mathcal{B}_i)\cdot \nabla \textit{exp} \Bigl( - \frac{\lVert \widetilde{\theta}-\theta \rVert^2}{2\sigma^2} \Bigr)d \widetilde{\theta} \notag \\ 
    &= (2 \pi)^{-\frac{d}{2}} \int F_i(\theta + \delta, \mathcal{B}_i)\cdot \frac{\delta}{\sigma^2} \cdot \textit{exp} \Bigl( - \frac{\lVert \delta \rVert^2}{2\sigma^2} \Bigr) d \delta \notag \\
    &= \mathbb{E}_{\delta \sim \mathcal{N}(0, \sigma^2 I_d)} \Bigl[ \frac{\delta}{\sigma^2} \cdot F_i(\theta + \delta, \mathcal{B}_i) \Bigr] \notag \\
    &= \mathbb{E}_{\mu z_i \sim \mathcal{N}(0, \mu^2 I_d)} \Bigl[ \frac{z_i}{\mu} \cdot F_i(\theta + \mu z_i, \mathcal{B}_i) \Bigr].
\end{align}

By symmetry, we change $\delta$ to $-\delta$ and obtain
\begin{align}
    \nabla F_i(\theta, \mathcal{B}_i) &= -\mathbb{E}_{\mu z_i \sim \mathcal{N}(0, \mu^2 I_d)} \Bigl[ \frac{z_i}{\mu} \cdot F_i(\theta - \mu z_i, \mathcal{B}_i) \Bigr].
\end{align}

Further, we prove that with
\begin{align}
    \nabla F_i(\theta, \mathcal{B}_i) &= \frac{1}{2}\mathbb{E} \Bigl[ \frac{z_i}{\mu} \cdot F_i(\theta + \mu z_i, \mathcal{B}_i) \Bigr] - \frac{1}{2} \mathbb{E} \Bigl[ \frac{z_i}{\mu} \cdot F_i(\theta - \mu z_i, \mathcal{B}_i) \Bigr] \notag \\
    &= \mathbb{E} \Bigl[ \frac{z_i}{2 \mu} \widetilde{\nabla} F_i(\theta, z_i, \mathcal{B}_i, \mu )\Bigr].
\end{align}

Finally, we adopt Eq. \eqref{eq:pre_fl_origin} and get
Lemma \ref{lemma:unbiased}, i.e.,

\begin{equation}
\nabla f_i(\theta) = \mathbb{E}[\nabla F_i(\theta, \mathcal{B}_i)]  = \mathbb{E}[ \widetilde{\nabla} F_i(\theta, z_i, \mathcal{B}_i, \mu) ].
\end{equation} \qed

\subsection{Proof of Lemma \ref{lemma:dimension_free}}
\label{appendix:lemma_dimension}

Malladi et al. \cite{malladi2023fine} present a step-wise learning rate decay corollary that is independent of the dimensionality parameter $d$ and is solely related to the low effective rank $r$. We formalize this result as follows:

\begin{lemma}[Step-Wise Learning Rate Decay]
\label{lemma:dimension_free_origin}
Assuming the Hessian matrix in terms of $\theta$ exhibits a local effective rank of $r$, and there exists a constant $\gamma = \frac{dr +d-2}{n(d+2)} =  \Theta(r / n)$, the expected decrease in the loss can be bounded as follows:
    \begin{align}
        \label{eq:each_step_low_rank_origin}
        \mathbb{E}\bigl[ f(\theta^{t+1}) \bigr] \leq f(\theta^{t}) - \frac{1}{\gamma} \eta \lVert \nabla f(\theta^{t}) \rVert^2 + \frac{1}{2} \eta^2 L \frac{1}{\gamma} \mathbb{E}\bigl[ {\lVert \nabla{F(\theta, \mathcal{B})} \rVert}^2 \bigr], 
    \end{align}
where $n$ denotes the number of randomizations.
\end{lemma}
It is important to note that the last term in Eq. \eqref{eq:each_step_low_rank_origin} represents the squared norm of the true gradient, which is inconsistent with the zeroth-order estimation method used in our approach. Therefore, a transformation is necessary to align it with the \ours algorithm. Besides, Malladi et al. detail the relationship between the true gradient norm and the zeroth-order gradient estimator norm, which we restate as follows:
\begin{lemma}[Gradient Estimator Norm Relationship]
\label{lemma:diff_estimator}
The squared norm of the gradient estimated by the MeZO is given by
\begin{equation}
\label{eq:diff_estimator}
\mathbb{E}\Bigl[ {\lVert \nabla{F(\theta, \mathcal{B})} \rVert}^2 \Bigr] = \frac{n}{d+n-1} \mathbb{E}\Bigl[ {\lVert \widetilde{\nabla}{F(\theta, z, \mathcal{B}, \mu)} \rVert}^2 \Bigr]. 
\end{equation}
\end{lemma}

By substituting Eq. \eqref{eq:diff_estimator} into Eq. \eqref{eq:each_step_low_rank_origin}, we obtain Lemma \ref{lemma:dimension_free}. \qed

\section{Full Proof of Theorems}
\label{appendix:theorem}

\subsection{Proof of Theorem \ref{thm:iid}}
\label{appendix:thm_iid}
For the term $T_1$:
\begin{align}
    \label{eq:t1_iid_app}
    T_1 &= \mathbb{E}_t {\Bigg\lVert \frac{1}{N} \sum_{i=1}^N \nabla f_i(\theta^{t}) \Bigg\rVert}^2 \notag \\
    &= \mathbb{E}_t {\Bigg\lVert \frac{1}{N} \sum_{i=1}^N \Bigl[ \nabla f_i(\theta^{t}) - \nabla f(\theta^{t}) + \nabla f(\theta^{t}) \Bigr]   \Bigg\rVert}^2 \notag \\
    &\leq 2\mathbb{E}_t {\Bigg\lVert \frac{1}{N} \sum_{i=1}^N \Bigl[ \nabla f_i(\theta^{t}) - \nabla f(\theta^{t}) \Bigr]   \Bigg\rVert}^2 + 2\mathbb{E}_t {\Bigg\lVert \frac{1}{N} \sum_{i=1}^N  \nabla f(\theta^{t}) \Bigg\rVert}^2 \notag \\
    &\leq \frac{2}{N^2} \sum_{i=1}^N \mathbb{E}_t {\Big\lVert \nabla f_i(\theta^{t}) - \nabla f(\theta^{t}) \Big\rVert}^2 + 2\mathbb{E}_t {\big\lVert \nabla f(\theta^{t}) \big\rVert}^2 \notag \\
    &= 2 \mathbb{E}_t {\big\lVert \nabla f(\theta^{t}) \big\rVert}^2,
\end{align}
where the first inequality follows from the Cauchy-Schwarz inequality, the second inequality follows from Jensen's inequality, and the third equality is derived from Assumption \ref{assumption:local_global_iid}.

For the term $T_2$, by applying Jensen's inequality, we obtain:
\begin{equation}
    T_2 \leq \frac{1}{N^2} \sum_{i=1}^N \sum_{k=1}^H \mathbb{E}_t {\Big\lVert  e_i^{(t,k)} \Big\rVert}^2,
\end{equation}
where $e_i^{(t,k)}$ represents the gradient estimator defined in Eq. \eqref{eq:pre_each_step_low_rank_estimator}. Substituting the estimator into the above expression and simplifying it, we obtain the following inequality. For brevity, let $\theta_i^{'}$ denote $\theta_i^{(t,k)}$, $z_i^{'}$ denote $z_i^{(t,k)}$, and $\mathcal{B}i^{'}$ denote $\mathcal{B}i^{(t,k)}$:

\begin{align}
\label{eq:t2_apart_iid_app}
    T_2 &\leq \frac{1}{N^2} \sum_{i=1}^N \sum_{k=1}^H \mathbb{E}_t {\Bigg\lVert  \frac{z_i^{'}}{2\mu}(F_i(\theta_i^{'}+\mu z_i^{'}, \mathcal{B}_i^{'})-F_i(\theta_i^{'}-\mu z_i^{'}, \mathcal{B}_i^{'}) \Bigg\rVert}^2 \notag \\
    &= \frac{1}{N^2} \sum_{i=1}^N \sum_{k=1}^H \mathbb{E}_t \Bigg\lVert  \frac{z_i^{'}}{2\mu}\Bigl( F_i(\theta_i^{'}+\mu z_i^{'}, \mathcal{B}_i^{'}) -  F_i(\theta_i^{'}, \mathcal{B}_i^{'}) \Bigr) \notag \\
    &\hspace{3.5cm} + \Bigl( F_i(\theta_i^{'}, \mathcal{B}_i^{'}) - F_i(\theta_i^{'}-\mu z_i^{'}, \mathcal{B}_i^{'}) \Bigr) \Bigg\rVert^2 \notag \\
    & \leq \frac{2}{N^2} \sum_{i=1}^N \sum_{k=1}^H \mathbb{E}_t \Bigg\lVert  \frac{z_i^{'}}{2\mu}\Bigl( F_i(\theta_i^{'}+\mu z_i^{'}, \mathcal{B}_i^{'}) -  F_i(\theta_i^{'}, \mathcal{B}_i^{'}) \Bigr) \Bigg\rVert^2 \notag \\
    & \hspace{0.5cm} + \frac{2}{N^2} \sum_{i=1}^N \sum_{k=1}^H \mathbb{E}_t \Bigg\lVert  \frac{z_i^{'}}{2\mu}\Bigl( F_i(\theta_i^{'}, \mathcal{B}_i^{'}) - F_i(\theta_i^{'}-\mu z_i^{'}, \mathcal{B}_i^{'}) \Bigr) \Bigg\rVert^2,
\end{align}
where the inequality follows from the fact that $(\lVert a + b \rVert)^2 \leq 2(\lVert a \rVert)^2 + 2(\lVert b \rVert)^2$. Due to the symmetry of the function $F_i$ when perturbed with Gaussian-distributed $z_i^{'}$, both terms on the right-hand side are equivalent.
Hence $T_2$ can be transformed into 

\begin{align}
\label{eq:t2_trans_iid_app}
    T_2 &\leq \frac{1}{N^2} \sum_{i=1}^N \sum_{k=1}^H \mathbb{E}_t \Bigg\lVert  \frac{z_i^{'}}{\mu}\Bigl( F_i(\theta_i^{(t,k)}+\mu z_i^{'}, \mathcal{B}_i^{'}) -  F_i(\theta_i^{'}, \mathcal{B}_i^{'}) \Bigr) \Bigg\rVert^2 \notag \\
    &= \frac{1}{N^2 \cdot d^2} \sum_{i=1}^N \sum_{k=1}^H \mathbb{E}_t \Bigg\lVert  \frac{d \cdot z_i^{'}}{\mu}\Bigl( F_i(\theta_i^{'}+\mu z_i^{'}, \mathcal{B}_i^{'}) -  F_i(\theta_i^{'}, \mathcal{B}_i^{'}) \Bigr) \Bigg\rVert^2 \notag \\
    &\leq \frac{1}{N^2 \cdot d^2} \sum_{i=1}^N \sum_{k=1}^H \Biggl[ 2d\cdot \mathbb{E}_t \Big\lVert \nabla F_i(\theta_i^{'}, \mathcal{B}_i^{'})\Big\rVert^2 + \frac{\mu^2}{2} L^2 d^2 \Biggr]\notag \\
    &\leq \frac{1}{N^2 \cdot d^2} \sum_{i=1}^N \sum_{k=1}^H \Biggl[ 2 c_g d\cdot \mathbb{E}_t \Big\lVert \nabla f_i(\theta_i^{'})\Big\rVert^2 + 2 d \sigma_g^2+ \frac{\mu^2}{2} L^2 d^2 \Biggr],
\end{align}
where the first inequality follows Lemma 4.1 from \cite{gao2018information}, and the second inequality follows the Assumption \ref{assumption:minibatch}. For the expectation term, we have 

\begin{align}
\label{eq:t2_fi_iid_app}
    \mathbb{E}_t \Big\lVert \nabla f_i(\theta_i^{(t,k)})\Big\rVert^2 &= \mathbb{E}_t \Big\lVert \nabla f_i(\theta_i^{(t,k)}) \mp \nabla f_i(\theta^t) \mp \nabla f(\theta_i^t) \Big\rVert^2 \notag \\
    & \leq 3 L^2 \mathbb{E}_t \Big\lVert \theta_i^{(t,k)} - \theta^t \Big\rVert^2 + 3\mathbb{E}_t \Big\lVert \nabla f(\theta^t) \Big\rVert^2,
\end{align}
where the inequality is due to the Cauchy-Schwartz inequality, $L$-smooth property and Assumption \ref{assumption:local_global_iid}. 
By taking Eq. \eqref{eq:t2_fi_iid_app} into Eq. \eqref{eq:t2_trans_iid_app}, finally we can bound $T_2$ as:

\begin{align}
\label{eq:t2_final_iid_app}
    T_2 \leq \frac{6 c_g L^2}{N^2d} &\sum_{i=1}^N \sum_{k=1}^H \mathbb{E}_t \Big\lVert \theta_i^{(t,k)} - \theta^t \Big\rVert^2  \notag \\
    &+ \frac{6 c_gH}{Nd} \mathbb{E}_t \Big\lVert \nabla f(\theta^t) \Big\rVert^2  + \frac{2 H \sigma_g^2}{Nd} + \frac{\mu^2 H L^2}{2N}.
\end{align}

After combining Eq. \eqref{eq:each_step_start_with}, Eq. \eqref{eq:t1_iid_app} and Eq. \eqref{eq:t2_final_iid_app}, we have

\begin{align}
\label{eq:each_step_t1t2_iid_app}
    \mathbb{E}_t &\bigl[ f(\theta^{t+1}) \bigr] \leq f(\theta^{t}) - \frac{2}{\gamma} \eta \mathbb{E}_t \Big\lVert \nabla f(\theta^{t}) \Big\rVert^2 + \frac{3 c_g \zeta \eta^2 H L}{Nd} \mathbb{E}_t \Big\lVert \nabla f(\theta^t) \Big\rVert^2 \notag \\
    &+ \frac{3 c_g \zeta \eta^2 L^3}{N^2 d} \sum_{i=1}^N \sum_{k=1}^H \mathbb{E}_t \Big\lVert \theta_i^{(t,k)} - \theta^t \Big\rVert^2 + \frac{\sigma_g^2 \zeta \eta^2 H  L}{Nd} + \frac{\zeta \eta^2 \mu^2 H L^3 }{4N} \notag \\
    &= f(\theta^{t}) + \Biggl( \frac{3 c_g \zeta \eta^2 H L}{Nd} - \frac{2}{\gamma} \cdot \eta \Biggr) \mathbb{E}_t \Big\lVert \nabla f(\theta^{t}) \Big\rVert^2  \notag \\
    &+ \frac{3 c_g \zeta \eta^2 L^3}{N^2 d} \sum_{i=1}^N \sum_{k=1}^H \mathbb{E}_t \Big\lVert \theta_i^{(t,k)} - \theta^t \Big\rVert^2 + \frac{\sigma_g^2 \zeta \eta^2 H L}{Nd} + \frac{\zeta \eta^2 \mu^2 H L^3 }{4N}.
\end{align}

Next we need to bound $\mathbb{E}_t \Big\lVert \theta_i^{(t,k)} - \theta^t \Big\rVert^2$ and simplify Eq. \eqref{eq:each_step_t1t2_iid_app}. Specifically, by denoting $\frac{1}{N} \sum_{i=1}^N \mathbb{E}_t^{k-1}\lVert \theta_i^{(t,k)} - \theta^t \rVert^2$ as $s^{(t,k)}$, we have

\begin{align}
\label{eq:lemma2_start_with_iid_app}
    s^{(t,\tau)} &= \eta^2 \frac{1}{N} \sum_{i=1}^N \mathbb{E}_t^{\tau-1} \Bigg\lVert \sum_{k=1}^{\tau} e_i^{(t,k)} \Bigg\rVert^2 \notag \\
    & \leq \tau  \eta^2 \sum_{k=1}^{\tau} \frac{1}{N} \sum_{i=1}^N \mathbb{E}_t^k \Big\lVert  e_i^{(t,k)} \Big\rVert^2.
\end{align}

By combing Eq. \eqref{eq:t2_trans_iid_app}, Eq. \eqref{eq:t2_fi_iid_app} and Eq. \eqref{eq:lemma2_start_with_iid_app}, we have

\begin{align}
\label{eq:lemma2_2_iid_app}
    s^{(t,\tau)} \leq 6 c_g d L^2 \tau \eta^2 \sum_{k=1}^{\tau}  s^{(t,k)} + 6& c_g d \tau^2 \eta^2 \mathbb{E}_t\big\lVert \nabla f(\theta^t) \big\rVert^2 \notag \\
    &+ 2d\sigma_g^2 \tau^2 \eta^2 + \frac{\mu^2 L^2 d^2 \tau^2 \eta^2}{2}.
\end{align}

By taking summation over $\tau$ from 2 to $H$, and utilizing the property of arithmetic sequence, we obtain

\begin{align}
\label{eq:lemma2_3_iid_app}
    \sum_{\tau=2}^H s^{(t,\tau)} &\leq 6 c_g d L^2  \eta^2 \sum_{\tau=2}^H \tau \sum_{k=1}^{\tau}  s^{(t,k)} + C_1 \notag \\
    & \leq  3 c_g d H^2 L^2  \eta^2 \sum_{k=1}^{H}  s^{(t,k)} + C_1,  \\
    \textit{where}\hspace{0.1cm} C_1 = 2 c_g d H^3 &\eta^2 \mathbb{E}_t \big\lVert \nabla f(\theta^t) \big\rVert^2 + \frac{2}{3}d\sigma_g^2 H^3 \eta^2 + \frac{\mu^2 L^2 d^2 H^3 \eta^2}{6}. \notag
\end{align}

As $s^{(t,1)}=0$, after rearranging Eq. \eqref{eq:lemma2_3_iid_app}, we have  
\begin{align}
\label{eq:lemma2_4_iid_app}
    (1-3 c_g d H^2 L^2  \eta^2) \sum_{k=1}^{H}  s^{(t,k)} \leq C_1.
\end{align}

For simplification, here we denote $(1-3 c_g d H^2 L^2  \eta^2)$ as $C_0$. When $\eta \leq \frac{1}{3 H L \sqrt{c_g d}}$, $C_0 \geq \frac{2}{3}$. Under this condition, we have:
\begin{align}
\label{eq:lemma2_final_iid_app}
    \frac{1}{N} \sum_{i=1}^N \sum_{k=1}^H \mathbb{E}_t \Big\lVert \theta_i^{(t,k)} - \theta^t \Big\rVert^2 &= \sum_{k=1}^{H}  s^{(t,k)} \leq \frac{C_1}{C_0}  \notag \\
    = \frac{1}{C_0} \Biggl( 2 c_g d H^3 \eta^2 \mathbb{E}_t\big\lVert \nabla f(\theta^t) \big\rVert^2 &+ \frac{2}{3}d\sigma_g^2 H^3 \eta^2 + \frac{\mu^2 L^2 d^2 H^3 \eta^2}{6} \Biggr) \notag \\
    \leq  3 c_g d H^3 \eta^2 \mathbb{E}_t\big\lVert \nabla f(\theta^t) \big\rVert^2 &+ d \sigma_g^2 H^3 \eta^2 + \frac{\mu^2 L^2 d^2 H^3 \eta^2}{4}.
\end{align}

Taking Eq. \eqref{eq:lemma2_final_iid_app} into Eq. \eqref{eq:each_step_t1t2_iid_app}, we obtain the final result of loss descent of each step:
\begin{align}
\label{eq:each_step_final_iid_app}
    &\mathbb{E}_t \bigl[ f(\theta^{t+1}) \bigr] \leq f(\theta^{t}) + \Biggl( \frac{3 c_g \zeta \eta^2 H L}{Nd} - \frac{2}{\gamma} \cdot \eta \Biggr) \mathbb{E}_t \Big\lVert \nabla f(\theta^{t}) \Big\rVert^2 \notag \\
    &\quad+ \frac{3 c_g \zeta \eta^2 L^3}{Nd} \cdot \frac{C_1}{C_0} + \frac{\sigma_g^2 \zeta \eta^2 H L}{Nd} + \frac{\zeta \eta^2 \mu^2 H L^3 }{4N} \notag \\
    & \leq f(\theta^{t}) + \Biggl( \frac{3 c_g \zeta \eta^2 H L}{Nd} - \frac{2}{\gamma} \cdot \eta+ \frac{9 c_g^2 \zeta \eta^4 H^3 L^3}{N}  \Biggr) \mathbb{E}_t \Big\lVert \nabla f(\theta^{t}) \Big\rVert^2 \notag \\
    & \quad + \frac{3 c_g \sigma_g^2 \zeta \eta^4 H^3 L^3}{N} + \frac{3 c_g d \zeta \eta^4 \mu^2 H^3 L^5 }{4N}\notag + \frac{\sigma_g^2 \zeta \eta^2 H L}{Nd} + \frac{\zeta \eta^2 \mu^2 H L^3 }{4N} \\
    & \leq f(\theta^{t}) + \Biggl( \frac{3 c_g \zeta \eta^2 H L}{Nd} + 3 c_g d \eta^2 H^2 L^2 \frac{3 c_g \zeta \eta^2 H L}{Nd} \Biggr) \mathbb{E}_t \Big\lVert \nabla f(\theta^{t}) \Big\rVert^2 \notag \\
    & \quad -\frac{2\eta}{\gamma} \mathbb{E}_t \Big\lVert \nabla f(\theta^{t}) \Big\rVert^2 + 3 c_g d \eta^2 H^2 L^2 \cdot \frac{\sigma_g^2 \zeta \eta^2 H L}{Nd}  \notag \\
    & \quad + 3 c_g d \eta^2 H^2 L^2 \cdot \frac{\zeta \eta^2 \mu^2 H L^3 }{4N} + \frac{\sigma_g^2 \zeta \eta^2 H L}{Nd} + \frac{\zeta \eta^2 \mu^2 H L^3 }{4N}\notag \\
    & \leq f(\theta^{t}) + \Biggl( (1+ 3 c_g d H^2 L^2 \eta^2) \frac{3 c_g \zeta \eta^2 H L}{Nd} - \frac{2}{\gamma} \cdot \eta \Biggr) \mathbb{E}_t \Big\lVert \nabla f(\theta^{t}) \Big\rVert^2 \notag \\
    & \quad + (1+ 3 c_g d H^2 L^2 \eta^2) \frac{\sigma_g^2 \zeta \eta^2 H L}{Nd} + (1+ 3 c_g d H^2 L^2 \eta^2) \frac{\zeta \eta^2 \mu^2 H L^3 }{4N}.
\end{align}

With the condition $\eta \leq \frac{1}{3 H L \sqrt{c_g d}}$, we have $(1+ 3 c_g d H^2 L^2 \eta^2) \leq 2$.  Then taking the condition $\eta \leq \frac{N}{3 c_g H L}$ and $\eta \leq \frac{1}{H^2}$, we can transform Eq. \eqref{eq:each_step_final_iid_app} into the result of Theorem \ref{thm:iid} as:
\begin{align}
\label{eq:each_step_final_2_iid_app}
    \mathbb{E}_t \bigl[ f(\theta^{t+1}) \bigr]  \leq f(\theta^{t}) + 2 \Biggl( \frac{\zeta}{d} \eta &- \frac{1}{\gamma} \eta \Biggr) \mathbb{E}_t \Big\lVert \nabla f(\theta^{t}) \Big\rVert^2 \notag \\
    & \qquad + \frac{2 \sigma_g^2 \zeta \eta L}{NHd} + \frac{\zeta \eta \mu^2 L^3 }{2NH}.
\end{align} \qed

\subsection{Proof of Corollary \ref{col:iid}}
\label{appendix:col_iid}
To get the result of global convergence, we rearrange Eq. \eqref{eq:each_step_final_2_iid_app} and similarly bound $\eta$ as $\eta \leq \frac{1}{H^2}$. For simplicity, we denotes $\frac{d-\zeta \gamma}{d \gamma}$ as $\Gamma$ and have
\begin{align}
\label{eq:each_step_final_3_iid_app}
    2\eta \Gamma \mathbb{E}_t \Big\lVert \nabla f(\theta^{t}) \Big\rVert^2 &\leq f(\theta^{t}) - \mathbb{E}_t \bigl[ f(\theta^{t+1}) \bigr]  + \frac{2 \sigma_g^2 \zeta L}{NHd} \eta + \frac{\zeta \mu^2 L^3 }{2NH} \eta \notag \\
    \mathbb{E}_t \Big\lVert \nabla f(\theta^{t}) \Big\rVert^2 &\leq \frac{f(\theta^{t}) - \mathbb{E}_t \bigl[ f(\theta^{t+1})\bigr]}{2 \eta \Gamma} + \frac{\sigma_g^2 \zeta L}{\Gamma NHd} + \frac{\zeta \mu^2 L^3 }{4\Gamma NH}.
\end{align}

Notice that $\gamma = \Theta(r)$ and $\zeta = \Theta(\frac{1}{rd})$, since parameter $d$ is a large number, $d-\zeta \gamma = d- \Theta(\frac{1}{d}) = d$, hence the dominant term of $\Gamma$ is $\frac{1}{\gamma}$, which follows $\Gamma = \Theta(\frac{1}{r})$. Then simultaneously summing over $T$ rounds on both sides and taking the average:
\begin{align}
\label{eq:each_step_final_4_iid_app}
    \frac{1}{T} \sum_{t=0}^T \mathbb{E}_t \Big\lVert \nabla f(\theta^{t}) \Big\rVert^2 \leq \frac{f(\theta^{0}) - \mathbb{E}_t \bigl[ f(\theta^{T})\bigr]}{2 \eta T \Gamma} + \frac{\sigma_g^2 \zeta L}{\Gamma NHd} + \frac{\zeta \mu^2 L^3 }{4\Gamma NH}.
\end{align}
\qed

\subsection{Proof of Theorem \ref{thm:noniid}}
\label{appendix:thm_noniid}

Starting from Eq. \eqref{eq:each_step_start_with}, under the assumption of global-local dissimilarity (Assumption \ref{assumption:local_global_noniid}), $T_1$ becomes:

\begin{align}
    \label{eq:t1_noniid_app}
    T_1 &= \mathbb{E}_t {\Bigg\lVert \frac{1}{N} \sum_{i=1}^N \nabla f_i(\theta^{t}) \Bigg\rVert}^2 \notag \\
    &= \mathbb{E}_t {\Bigg\lVert \frac{1}{N} \sum_{i=1}^N \Bigl[ \nabla f_i(\theta^{t}) - \nabla f(\theta^{t}) + \nabla f(\theta^{t}) \Bigr]   \Bigg\rVert}^2 \notag \\
    &\leq 2\mathbb{E}_t {\Bigg\lVert \frac{1}{N} \sum_{i=1}^N \Bigl[ \nabla f_i(\theta^{t}) - \nabla f(\theta^{t}) \Bigr]   \Bigg\rVert}^2 + 2\mathbb{E}_t {\Bigg\lVert \frac{1}{N} \sum_{i=1}^N  \nabla f(\theta^{t}) \Bigg\rVert}^2 \notag \\
    &\leq \frac{2}{N^2} \sum_{i=1}^N \mathbb{E}_t {\Big\lVert \nabla f_i(\theta^{t}) - \nabla f(\theta^{t}) \Big\rVert}^2 + 2\mathbb{E}_t {\big\lVert \nabla f(\theta^{t}) \big\rVert}^2 \notag \\
    &\leq \frac{2}{N^2} \sum_{i=1}^N \mathbb{E}_t \Bigl[ c_h \big\lVert \nabla f(\theta^{t}) \big\rVert + \sigma_h^2
    \Bigr] + 2 \mathbb{E}_t {\big\lVert \nabla f(\theta^{t}) \big\rVert}^2 \notag \\
    &= \frac{2(N+c_h)}{N} \mathbb{E}_t {\big\lVert \nabla f(\theta^{t}) \big\rVert}^2 + \frac{2 \sigma_h^2}{N},
\end{align}
where the first inequality follows Cauchy-Schwartz, the second inequality follows Jensen's inequality, and the third inequality follows Assumption \ref{assumption:local_global_noniid}.

Then we begin to bound the term $T_2$. Taking the result from Eq. \eqref{eq:t2_trans_iid_app}, we have:
\begin{equation}
\label{eq:t2_trans_noniid_app}
    T_2 \leq \frac{1}{N^2 \cdot d^2} \sum_{i=1}^N \sum_{k=1}^H \Biggl[ 2 c_g d\cdot \mathbb{E}_t \Big\lVert \nabla f_i(\theta_i^{(t,k)})\Big\rVert^2 + 2 d \sigma_g^2+ \frac{\mu^2}{2} L^2 d^2 \Biggr],
\end{equation}
where the first inequality follows Lemma 4.1 from \cite{gao2018information}, and the second inequality follows the Assumption \ref{assumption:minibatch}. The term $\mathbb{E}_t \Big\lVert \nabla f_i(\theta_i^{(t,k)})\Big\rVert^2$ in Eq. \eqref{eq:t2_trans_noniid_app} can be bounded as:
\begin{align}
\label{eq:t2_fi_noniid_app}
    \mathbb{E}_t \Big\lVert \nabla &f_i(\theta_i^{(t,k)})\Big\rVert^2 = \mathbb{E}_t \Big\lVert \nabla f_i(\theta_i^{(t,k)}) \mp \nabla f_i(\theta^t) \mp \nabla f(\theta_i^t) \Big\rVert^2 \notag \\
    & \leq 3 L^2 \mathbb{E}_t \Big\lVert \theta_i^{(t,k)} - \theta^t \Big\rVert^2 + 3(c_h+1)\mathbb{E}_t \Big\lVert \nabla f(\theta^t) \Big\rVert^2 + 3\sigma_h^2,
\end{align}
where the inequality follows the Cauchy-Schwartz, $L$-smooth and Assumption \ref{assumption:local_global_noniid}. 
By applying Eq. \eqref{eq:t2_fi_noniid_app} into Eq. \eqref{eq:t2_trans_noniid_app}, we obtain:
\begin{align}
\label{eq:t2_final_noniid_app}
    T_2 &\leq \frac{6 c_g L^2}{N^2d} \sum_{i=1}^N \sum_{k=1}^H \mathbb{E}_t \Big\lVert \theta_i^{(t,k)} - \theta^t \Big\rVert^2 + \frac{6 c_g (c_h+N)H}{N d} \mathbb{E}_t \Big\lVert \nabla f(\theta^t) \Big\rVert^2  \notag \\
    & \hspace{3cm} + \frac{6 c_g \sigma_h^2 H}{Nd}+ \frac{2 H \sigma_g^2}{Nd} + \frac{\mu^2 H L^2}{2N}.
\end{align}

Combining Eq. \eqref{eq:each_step_start_with}, Eq. \eqref{eq:t1_noniid_app} and Eq. \eqref{eq:t2_final_noniid_app}, we have:
\begin{align}
\label{eq:each_step_t1t2_noniid_app}
    &\mathbb{E}_t \bigl[ f(\theta^{t+1}) \bigr] \leq f(\theta^{t}) - \frac{2}{\gamma N} (N + c_h) \eta \mathbb{E}_t \Big\lVert \nabla f(\theta^{t}) \Big\rVert^2 - \frac{2}{\gamma N} \sigma_h^2 \eta \notag \\
    &\quad+ \frac{3c_g(c_h+N) \zeta \eta^2 H L}{Nd} \mathbb{E}_t \Big\lVert \nabla f(\theta^t) \Big\rVert^2 + \frac{3 c_g \sigma_h^2 \zeta \eta^2 H L}{Nd} \notag \\
    &\quad+ \frac{3 c_g \zeta \eta^2 L^3 }{N^2 d} \sum_{i=1}^N \sum_{k=1}^H \mathbb{E}_t \Big\lVert \theta_i^{(t,k)} - \theta^t \Big\rVert^2 + \frac{\sigma_g^2 \zeta \eta^2 H L}{Nd} + \frac{\zeta \eta^2 \mu^2 H L^3 }{4N} \notag \\
    &= f(\theta^{t}) + \Biggl( \frac{3 c_g (c_h+N) \zeta \eta^2 H L}{Nd} - \frac{2}{\gamma N} \cdot (N + c_h)\eta \Biggr) \mathbb{E}_t \Big\lVert \nabla f(\theta^{t}) \Big\rVert^2  \notag \\
    &\quad+ \frac{3 c_g \zeta \eta^2 L^3 }{N^2 d} \sum_{i=1}^N \sum_{k=1}^H \mathbb{E}_t \Big\lVert \theta_i^{(t,k)} - \theta^t \Big\rVert^2 + \frac{3 c_g \sigma_h^2 \zeta \eta^2 H L}{Nd} \notag \\
    &\quad+ \frac{\sigma_g^2 \zeta \eta^2 H L}{Nd} + \frac{\zeta \eta^2 \mu^2 H L^3 }{4N} - \frac{2}{\gamma N} \cdot \sigma_h^2 \eta.
\end{align}

Similar to Section \ref{appendix:thm_iid}, now we bound $\mathbb{E}_t \Big\lVert \theta_i^{(t,k)} - \theta^t \Big\rVert^2$ as follows:
\begin{align}
\label{eq:lemma2_start_with_noniid_app}
    s^{(t,\tau)} &= \eta^2 \frac{1}{N} \sum_{i=1}^N \mathbb{E}_t^{\tau-1} \Bigg\lVert \sum_{k=1}^{\tau} e_i^{(t,k)} \Bigg\rVert^2 \notag \\
    & \leq \tau  \eta^2 \sum_{k=1}^{\tau} \frac{1}{N} \sum_{i=1}^N \mathbb{E}_t^k \Big\lVert  e_i^{(t,k)} \Big\rVert^2,
\end{align}
where $s^{(t,k)}$ denotes $\frac{1}{N} \sum_{i=1}^N \mathbb{E}_t^{k-1}\lVert \theta_i^{(t,k)} - \theta^t \rVert^2$.

By combing Eq. \eqref{eq:t2_trans_noniid_app}, Eq. \eqref{eq:t2_fi_noniid_app} and Eq. \eqref{eq:lemma2_start_with_noniid_app}, we have
\begin{align}
\label{eq:lemma2_2_noniid_app}
    s^{(t,\tau)} &\leq 6 c_g d L^2 \tau \eta^2 \sum_{k=1}^{\tau}  s^{(t,k)} + 6 c_g d (c_h + 1) \tau^2 \eta^2 \mathbb{E}_t \big\lVert \nabla f(\theta^t) \big\rVert^2 \notag \\
    & \hspace{1.5cm} + 6c_g d \sigma_h^2 \tau^2 \eta^2 + 2d\sigma_g^2 \tau^2 \eta^2 + \frac{\mu^2 L^2 d^2 \tau^2 \eta^2}{2}.
\end{align}

By taking summation over $\tau$ from 2 to $H$, and utilizing the property of arithmetic sequence, we obtain
\begin{align}
\label{eq:lemma2_3_noniid_app}
    \sum_{\tau=2}^H s^{(t,\tau)} &\leq 6 c_g d L^2  \eta^2 \sum_{\tau=2}^H \tau \sum_{k=1}^{\tau}  s^{(t,k)} + C_2 \notag \\
    & \leq  3 c_g d H^2 L^2  \eta^2 \sum_{k=1}^{H}  s^{(t,k)} + C_2  \\
    \textit{where} \quad C_2 = 2 c_g &d (c_h + 1) H^3 \eta^2 \mathbb{E}_t \big\lVert \nabla f(\theta^t) \big\rVert^2 \notag \\
    + 2c_g &d \sigma_h^2 H^3 \eta^2 + \frac{2}{3}d\sigma_g^2 H^3 \eta^2 + \frac{\mu^2 L^2 d^2 H^3 \eta^2}{6}. \notag
\end{align}

Rearranging Eq. \eqref{eq:lemma2_3_noniid_app} and using $s^{(t,1)}=0$, we have 
\begin{align}
\label{eq:lemma2_4_noniid_app}
    (1-3 c_g d H^2 L^2  \eta^2) \sum_{k=1}^{H}  s^{(t,k)} \leq C_2.
\end{align}

Let $C_0$ be $(1-3 c_g d H^2 L^2  \eta^2)$. When $\eta \leq \frac{1}{3 H L \sqrt{c_g d}}$, $C_0 \geq \frac{2}{3}$. Under this condition, we have:
\begin{align}
\label{eq:lemma2_final_noniid_app}
    \frac{1}{N} \sum_{i=1}^N \sum_{k=1}^H \mathbb{E}_t \Big\lVert \theta_i^{(t,k)} - \theta^t \Big\rVert^2 &= \sum_{k=1}^{H}  s^{(t,k)} \leq \frac{C_2}{C_0}  \notag \\
    \leq  3 c_g (c_h + 1) d \eta^2 H^3 \mathbb{E}_t \big\lVert &\nabla f(\theta^t) \big\rVert^2 + 3c_g \sigma_h^2 d \eta^2 H^3 \notag \\
    &+ \sigma_g^2 d \eta^2 H^3 + \frac{d^2 \eta^2 \mu^2 H^3 L^2 }{4}.
\end{align}

Let $\widetilde{c}_h$ be $(c_h + N)$ and $\widetilde{\sigma}^2$ be $(3 c_g \sigma_h^2 + \sigma_g^2)$.
Applying Eq. \eqref{eq:lemma2_final_noniid_app} into Eq. \eqref{eq:each_step_t1t2_noniid_app}, we can obtain the final result of stepwise loss descent:
\begin{align}
\label{eq:each_step_final_noniid_app}
    &\mathbb{E}_t \bigl[ f(\theta^{t+1}) \bigr] \leq f(\theta^{t}) + \frac{3\zeta \eta^2 L  H c_g(c_h+N)}{Nd}  \mathbb{E}_t \Big\lVert \nabla f(\theta^{t}) \Big\rVert^2 \notag \\
    & \qquad - \frac{2}{\gamma N} \cdot \eta (N + c_h) \mathbb{E}_t \Big\lVert \nabla f(\theta^{t}) \Big\rVert^2 + \frac{3\zeta \eta^2 L^3 c_g}{N d} \cdot \frac{C_2}{C_0} \notag \\
    & \qquad  + \frac{3\zeta \eta^2 c_g \sigma_h^2 H L}{Nd} + \frac{\zeta \eta^2 H \sigma_g^2 L}{Nd} + \frac{\zeta \eta^2 \mu^2 H L^3 }{4N} - \frac{2}{\gamma N} \cdot \eta \sigma_h^2 \notag \\
    & \leq f(\theta^{t}) + \Biggl( \frac{3\zeta \eta^2 L  H c_g \widetilde{c}_h}{N d} - \frac{2\eta \widetilde{c}_h}{\gamma N}  + \frac{9 \zeta c_g^2 \widetilde{c}_h H^3 L^3 \eta^4}{N} \Biggr) \mathbb{E}_t \Big\lVert \nabla f(\theta^{t}) \Big\rVert^2 \notag \\
    & \qquad + \frac{3 c_g \widetilde{\sigma}^2 \zeta H^3 L^3 \eta^4}{N} + \frac{3 c_g \zeta d \mu^2 H^3 L^5 \eta^4}{4N} + \frac{\widetilde{\sigma}^2 \zeta H L \eta^2}{Nd} \notag \\
    & \qquad+ \frac{\zeta \eta^2 \mu^2 H L^3 }{4N} - \frac{2}{\gamma N} \cdot \eta \sigma_h^2 \notag \\
    & \leq f(\theta^{t}) + \Biggl( (1+ 3 c_g d H^2 L^2 \eta^2) \frac{3\zeta \eta^2 L  H c_g \widetilde{c}_h}{N d} \Biggr) \mathbb{E}_t \Big\lVert \nabla f(\theta^{t}) \Big\rVert^2 \notag \\
    & \qquad - \frac{2}{\gamma N}\eta \widetilde{c}_h \mathbb{E}_t \Big\lVert \nabla f(\theta^{t}) \Big\rVert^2 + 3 c_g d H^2 L^2 \eta^2 \frac{\widetilde{\sigma}^2}{N d}\zeta H L \eta^2 - \frac{2}{\gamma N} \eta \sigma_h^2 \notag \\
    & \qquad  + 3 c_g d H^2 L^2 \eta^2 \frac{\zeta \eta^2 \mu^2 H L^3 }{4N} + \frac{\widetilde{\sigma}^2}{N d} \zeta H L \eta^2 + \frac{\zeta \eta^2 \mu^2 H L^3 }{4 N}  \notag \\
    & \leq f(\theta^{t}) + \Biggl((1+ 3 c_g d H^2 L^2 \eta^2) \frac{3\zeta \eta^2 L  H c_g \widetilde{c}_h}{N d} - \frac{2\eta \widetilde{c}_h}{\gamma N}  \Biggr) \mathbb{E}_t \Big\lVert \nabla f(\theta^{t}) \Big\rVert^2 \notag \\
    & \qquad + (1+3 c_g d H^2 L^2 \eta^2) \cdot \frac{\widetilde{\sigma}^2}{N d}\zeta H L \eta^2 \notag \\
    & \qquad+ (1+ 3 c_g d H^2 L^2 \eta^2) \cdot \frac{\zeta \eta^2 \mu^2 H L^3 }{4N} - \frac{2}{\gamma N} \cdot \eta \sigma_h^2.
\end{align}

With the condition $\eta \leq \frac{1}{3 H L \sqrt{c_g d}}$, we have $(1+ 3 c_g d H^2 L^2 \eta^2) \leq 2$. Taking $\eta \leq \frac{N}{3 c_g H L}$, $\eta \leq \frac{1}{H^2}$, Eq. \eqref{eq:each_step_final_noniid_app} becomes:
\begin{align}
\label{eq:each_step_final_2_noniid}
    \mathbb{E}_t \bigl[ f(\theta^{t+1}) \bigr] &\leq  f(\theta^{t}) + 2\Biggl(\frac{\zeta}{d} - \frac{1}{\gamma N} \Biggr) \widetilde{c}_h \eta \mathbb{E}_t \Big\lVert \nabla f(\theta^{t}) \Big\rVert^2 \notag \\
    & \qquad + \frac{2 \widetilde{\sigma}^2 \zeta L \eta}{N H d} + \frac{\zeta \eta \mu^2 L^3 }{2 N H} - \frac{2}{\gamma N} \cdot \eta \sigma_h^2.
\end{align}
\qed

\subsection{Proof of Corollary \ref{col:iid}}
\label{appendix:col_noniid}
Denote $\widetilde{\Gamma} = \frac{d - N \gamma \zeta}{d \gamma N}$.
Rearranging Eq. \eqref{eq:each_step_final_2_noniid}, simultaneously summing over $T$ rounds on both sides and taking the average, we get the result:
\begin{align}
\label{eq:each_step_final_3_noniid}
    2 \widetilde{\Gamma} \widetilde{c}_h \eta  \mathbb{E}_t \Big\lVert \nabla f(\theta^{t}) \Big\rVert^2 &\leq  f(\theta^{t}) -\mathbb{E}_t \bigl[ f(\theta^{t+1}) \bigr]  \notag \\
    &\qquad + \frac{2 \widetilde{\sigma}^2 \zeta L \eta}{N H d} + \frac{\zeta \eta \mu^2 L^3 }{2 N H} - \frac{2}{\gamma N} \cdot \eta \sigma_h^2 \notag \\
    \mathbb{E}_t \Big\lVert \nabla f(\theta^{t}) \Big\rVert^2 &\leq  \frac{f(\theta^{t}) -\mathbb{E}_t \bigl[ f(\theta^{t+1}) \bigr]}{2 \widetilde{\Gamma} \widetilde{c}_h \eta}  \notag \\
    & \qquad + \frac{ \widetilde{\sigma}^2 \zeta L}{\widetilde{\Gamma} \widetilde{c}_h N H d} + \frac{\zeta \mu^2 L^3 }{\widetilde{\Gamma} \widetilde{c}_h 4 N H} - \frac{\sigma_h^2}{\widetilde{\Gamma} \widetilde{c}_h \gamma N} \notag \\
    \frac{1}{T} \sum_{t=0}^T \mathbb{E}_t \Big\lVert \nabla f(\theta^{t}) \Big\rVert^2 &\leq  \frac{f(\theta^{0}) -f^*}{2 \widetilde{\Gamma} \widetilde{c}_h \eta T}  + \frac{ \widetilde{\sigma}^2 \zeta L}{\widetilde{\Gamma} \widetilde{c}_h N H d} \notag \\
    & \qquad+ \frac{\zeta \mu^2 L^3 }{\widetilde{\Gamma} \widetilde{c}_h 4 N H} - \frac{\sigma_h^2}{\widetilde{\Gamma} \widetilde{c}_h \gamma N}.
\end{align}
\qed

\section{Implementation Details}
\label{appendix:exp_setup}
In this section, we provide the detailed implementations of our experiments. Some experimental settings have already been discussed in Section \ref{subsec:exp_setup} and will not be reiterated here.

\subsection{Datasets}
\label{appendix:exp_datasets}
\begin{table}[ht]
\caption{Datasets and Basic Information.}
\label{table:datasets}

\centering
\begin{tabular}{@{}ccc@{}}
\toprule
Name & \#Sample & Domain \\ \midrule
Fed-Alpaca & 52.0k & Generic Language \\
Fed-Dolly & 15.0k & Generic Language \\
Fed-GSM8K & 7.5k & CoT \\
Fed-CodeAlpaca & 8.0k & Code Generation \\ \bottomrule
\end{tabular}
\end{table}

We adopt several federated tuning datasets tailored for LLMs from \cite{kuang2023federatedscope-LLM}, with
different splitting strategies to simulate the heterogeneity typical of different federated learning (FL) scenarios, including a uniform distribution of data, a Dirichlet distribution of data, and a splitter based on meta-information. The proportion of test data for each dataset is 1\%.

\textbf{Fed-Dolly:}
This federated corpus dataset, derived from \textit{Databricks-dolly-15k} \cite{conover2023free}, comprises eight categories of NLP tasks: brainstorming, classification, closed QA, creative writing, general QA, information extraction, open QA, and summarization. We divide the training set into three subsets using a three-way split and assign each subset to a distinct client.

\textbf{Fed-GSM8K:}
Constructed from the \textit{GSM8K} dataset \cite{cobbe2021training}, this collection is aimed at mathematical fine-tuning and consists of 7.5K training problems alongside 1K test problems. By default, we partition the training set uniformly into three subsets and allocate each to a separate client.

\textbf{Fed-CodeAlpaca:}
This federated version of \textit{CodeAlpaca} \cite{chaudhary2023code} encompasses code samples in ten programming languages, including C, C\#, C++, Go, Java, PHP, Pascal, Python, Scala, and X86-64 Assembly. Due to the scarcity of X86-64 Assembly samples in the original corpus, we exclude them. The remaining samples are then divided into three subsets using a default three-way split, with one subset assigned to each client.

\textbf{Fed-Alpaca:}
The \textit{Alpaca} dataset \cite{taori2023stanford} is designed for LLM fine-tuning and features natural language questions and responses for a variety of NLP tasks, such as text generation, translation, and open QA. It spans various domains like math, text processing, and code generation. 

\subsection{Experimental Platforms}
We implement our approaches using PyTorch \cite{paszke2019pytorch} v1.10.1, coupled with PEFT v0.3.0 and the Transformers library \cite{wolf2020transformers} v4.29.2. Experiments with LLaMA-3B are conducted on a computing platform equipped with four NVIDIA A100 GPUs (40GB), with pre-trained LLMs loaded as 16-bit floating-point numbers.

\subsection{Default Implementation Settings}

Following the guidelines in \cite{kuang2023federatedscope-LLM,malladi2023fine}, all approaches perform local training with a batch size of 1 to minimize memory usage. 
In an effort to standardize the experimental conditions, both backpropagation (BP)-based methods and our proposed method \ours, train locally with specific learning rates: $\eta = 1 \times 10^{-5}$ for the \textit{Fed-Dolly} and \textit{Fed-Alpaca} datasets, $\eta = 2 \times 10^{-5}$ for the \textit{Fed-CodeAlpaca} dataset, and $\eta = 2.5 \times 10^{-5}$ for the \textit{Fed-GSM8K} dataset. The rank and alpha parameters for Low-Rank Adaptation (LoRA) adapters used by both BP-based optimization and \ours are set to 128 and 256, respectively. As per \cite{malladi2023fine}, the perturbation scale $\mu$ for \ours is set to $1 \times 10^{-3}$. 
Unless otherwise stated, in our training process, we employed the early stopping mechanism to prevent over-fitting and reduce unnecessary training time caused following previous works \cite{chen2022pflbench}. The training was stopped if there was no improvement in the validation loss for a predefined number of consecutive epochs, known as the patience parameter. We chose a patience of 30 epochs based on empirical evidence or prior studies. The best model was selected from the epoch with the lowest validation loss. 
Furthermore, apart from individual experiments with time constraints (experiments in Appendix \ref{appendix:exp_lr_require} and Appendix \ref{appendix:exp_impact_client_num}), we conducted three sets of experiments with randomly selected seeds for the same set of parameters, and calculated the mean as the line plot with a 90\% confidence interval as the error bar.

The influence of different hyper-parameters for \ours has been analyzed in Section \ref{sec:exp}. 

\section{Supplementary Experiments}
\label{appendix:exp_all}

\subsection{Communication Cost of \ours}
\label{appendix:communication_cost}

In Section~\ref{subsec:problem_formula}, we mention using LoRA~\cite{hu2022lora} to reduce the substantial communication overhead. Specifically, \ours transmits only LoRA parameters, contrasting with FedAvg's full parameters uploads. We theoretically prove this method's equivalence to full-parameter transmission, formally expressed as follows: 

\begin{align}
    \theta_i^{t+1,0} =\frac{1}{N} \sum_{i=1}^{N}\theta_i^{t,H} = \frac{1}{N} \sum_{i=1}^{N} \left[ \theta_i^{t+1,0} + \sum_{k=1}^{H} \nabla_{lora} \theta_i^{t,k} \right] \notag \\ 
    = \theta_i^{t,0} + \frac{1}{N} \sum_{i=1}^{N} \sum_{k=1}^{H} \nabla_{lora} \theta_i^{t,k}
\end{align}

Note that the left side represents the client's parameters with FedAvg, while the right side corresponds to \ours, and they are equivalent. By utilizing LoRA, \ours achieves the same effect with just 1.23\% of parameters in our setting, totaling 42,598,400. Specifically, each parameter occupies 2 bytes under fp16, full parameter transmission needs 6.39GB, while LoRA demands merely 80.45MB.

\subsection{Computational Cost of \ours}
\label{appendix:computational_cost}

FedMeZO's single perturbation per iteration for gradient estimation substantially lowers computational costs versus BP-based optimizers. We empirically validate it in terms of GPU memory usage and time efficiency. Table~\ref{tab:gpu_memory} illustrates GPU memory usage. After deducting the base model's usage (8697MiB) from the total one in Table~\ref{tab:gpu_memory}, we can observe that FedMeZO requires only 7.68\% ~ 13.36\% of the usage compared to BP-based FedAvg during training. For time efficiency, we sample the duration of ten training rounds across four task and present the results in Table~\ref{table:time_efficiency}. The results indicate that FedMeZO requires 91.35\% ~ 95.24\% of the time taken by BP-based methods. Coupled with the faster loss decline shown in Figure~\ref{fig:main_convergence_app}, FedMeZO demonstrates both a quicker training speed and higher computational efficiency.
\begin{table}[h]
\caption{The average time for training 10-rounds of BP-based FedAvg and \ours.}
\label{table:time_efficiency}
\begin{tabular}{@{}ccc@{}}
\toprule
Task & BP-based FedAvg (s) & \ours (s) \\ \midrule
Dolly-Meta & 117 & 108 \\
CodeAlpaca-LDA & 126 & 120 \\
GSM8K-IID & 104 & 95 \\
Alpaca-IID & 107 & 99 \\\bottomrule
\end{tabular}
\end{table}

\subsection{Evaluations with Common LLM's Metrics}
\label{appendix:exp_llm_metrics}

For a more comprehensive evaluation, we examine \ours with some commonly used metrics for LLMs evaluations. We conduct evaluations on Dolly with MMLU metrics~\cite{hendrycks2020measuring}, Code with OpenAI-HumanEval metrics~\cite{chen2021evaluating}, and GSM8K with CoT metrics\cite{cobbe2021training}. We evaluate \ours and BP-based FedAvg at model checkpoints of rounds 0, 100, 200 and present the results in Table~\ref{table:llm_metrics}.

Comparing with the results in round 0, we find that FedMeZO gains 35.48\%, 1.52\%, 2.3\% average improvements on GSM8K, Code, and Dolly respectively after fed-tuning. By contrast, BP-based FedAvg gains 29.03\%, 0\% and 0.38\% respectively. The results verified FedMeZO's effectiveness again: 
\begin{itemize}
    \item Fine-tuning LLMs with FedMeZO effectively improves their performance on specific tasks;
    \item FedMeZO gains better performance compared to BP-based FedAvg.
\end{itemize}

\begin{table}[h]
\caption{Evaluations of \ours with LLM's Metrics}\label{table:llm_metrics}
\centering
\setlength\tabcolsep{2pt}
\begin{tabular}{ccccccc}
\toprule
\multirow{2}{*}{\textbf{Rounds}}& \multicolumn{2}{c}{\textbf{Dolly (\%)}}& \multicolumn{2}{c}{\textbf{Code (\%)}}& \multicolumn{2}{c}{\textbf{GSM8K (\%)}} \\
\cmidrule(lr){2-3} \cmidrule(lr){4-5} \cmidrule(lr){6-7}
& \ours& BP& \ours& BP& \ours& BP \\

\midrule
\textbf{0}& 26& 26& 8.53 & 8.53& 3.41& 3.41\\
\textbf{100}& \textbf{26.6}& 26.1& \textbf{8.66}& 8.53& \textbf{4.32}& 4.25 \\
\textbf{200}& \textbf{26.3}& 26.2& 8.41& \textbf{8.53}& \textbf{4.62}& 4.40 \\

\bottomrule

\end{tabular}
\end{table}

\subsection{Conditions for the Learning Rate}
\label{appendix:exp_lr_require}

\begin{figure}[htbp]
    \centering
    \includegraphics[width=0.85\columnwidth]{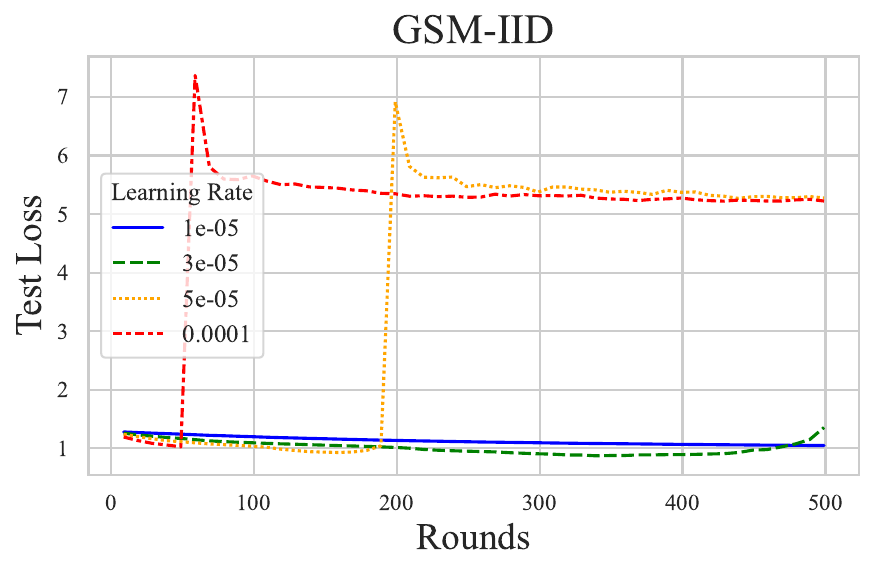}
    \caption{Phenomenon of loss surge due to larger learning rates.}
    \label{fig:lr_require}
\end{figure}

To validate the theory regarding the recommended learning rate settings mentioned in Section \ref{subsec:result_implications}, we conducted the following experiment. Firstly, we disabled the early-stop mechanism to observe the subsequent changes brought about by larger learning rates. Secondly, on the GSM-IID dataset, we sequentially selected four learning rates: $1\times10^{-5}$, $3\times10^{-5}$, $5\times10^{-5}$, and $1\times10^{-4}$, and conducted training for 500 rounds each. The final results are presented in Figure \ref{fig:lr_require}.

The results indicate that when the learning rate exceeds the range supported by theory, the loss function exhibits a sharp increase, and the larger the learning rate, the earlier this sharp increase occurs. 

It is noteworthy that our theory suggests that an optimal learning rate magnitude is anchored at $1 /\sqrt{d}$ in Section \ref{subsec:result_implications}. In our chosen model, LLaMA-3B, where $d$ can be set to $3\times10^9$, this gives $1 /\sqrt{d} = 1.826\times10^{-5}$, which is approximately the learning rate we aim to use. Exceeding this learning rate might lead to unexpected outcomes.

Therefore, this finding underscores the importance of adhering to theoretical guidelines for setting learning rates in order to avoid destabilizing the training process and ensuring a smooth convergence toward the optimal solution.

\subsection{The Impact of Heterogeneity on Convergence}
\label{appendix:exp_impact_hetero}

\begin{figure}[htbp]
    \centering
    \includegraphics[width=0.85\columnwidth]{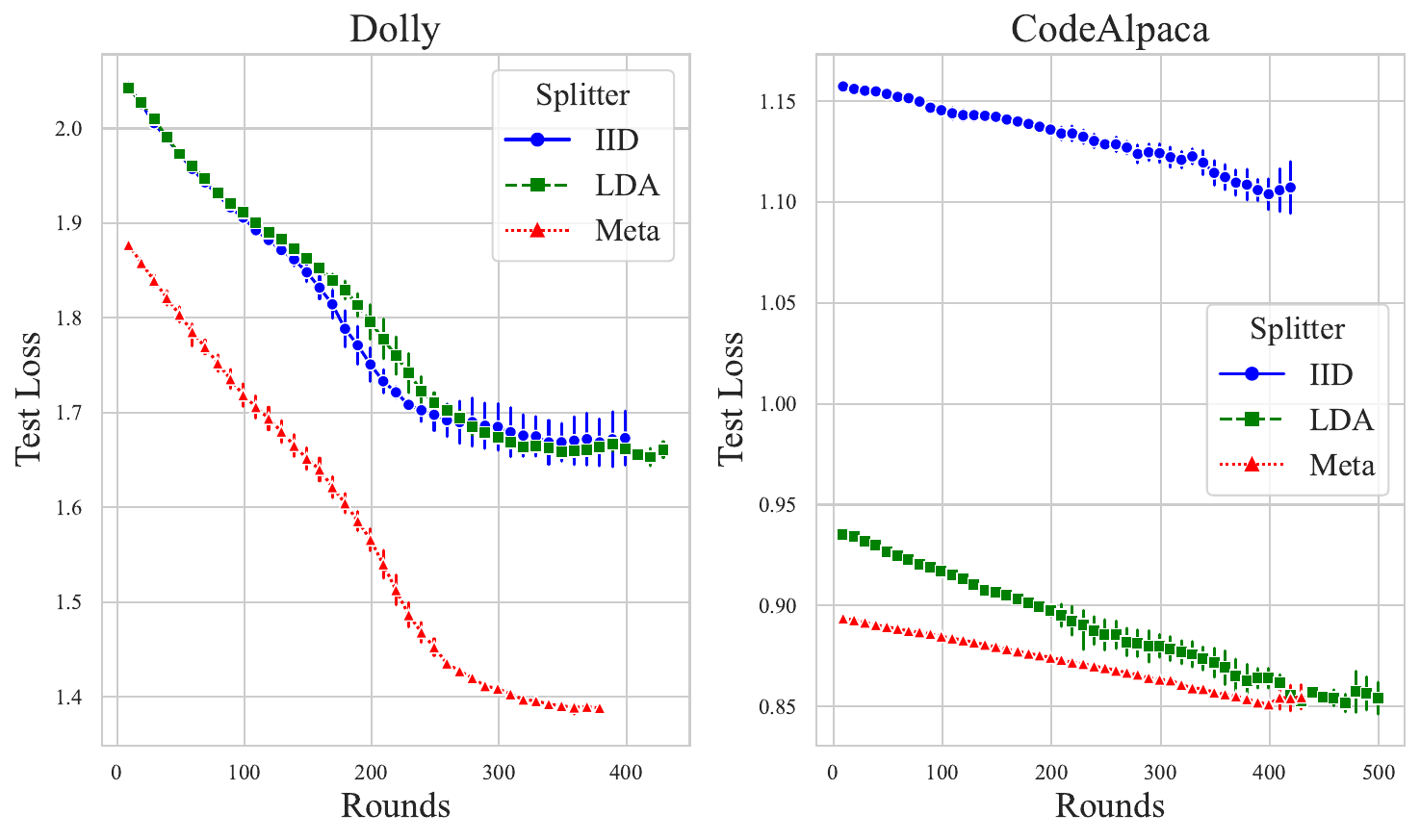}
    \caption{Effects of different splitters on the same dataset.}
    \label{fig:diff_split}
\end{figure}

We present the results of processing the same dataset with different splitters in Figure \ref{fig:diff_split}. It is observable that in the Dolly and CodeAlpaca datasets, the LDA and Meta splitters perform better than IID, with Meta being the best. Noting that the data classified by Meta and LDA are \noniid, this indicates that higher data heterogeneity is more conducive to model convergence.

\subsection{The Impact of Client Number}
\label{appendix:exp_impact_client_num}

\begin{figure}[htbp]
    \centering
    \includegraphics[width=0.85\columnwidth]{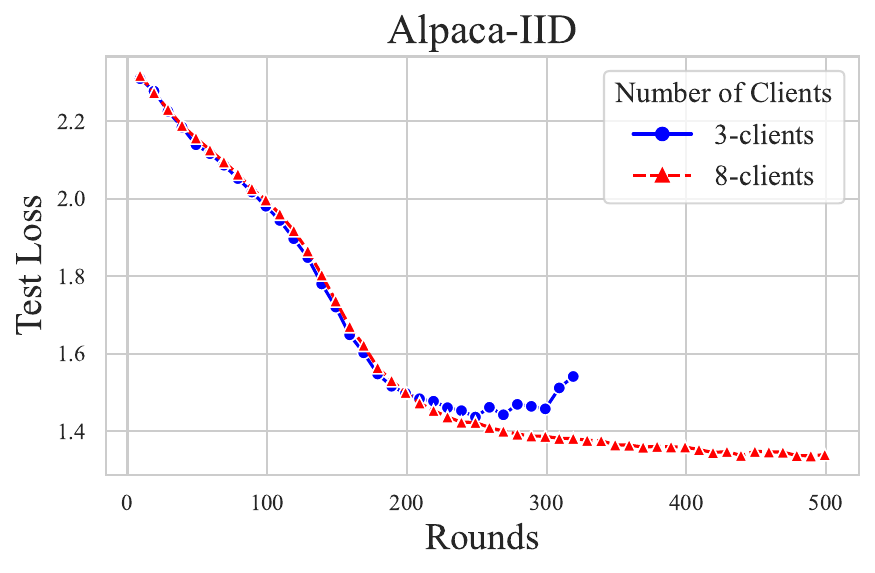}
    \caption{Effects of the number of clients.}
    \label{fig:diff_n}
\end{figure}

In Figure \ref{fig:diff_n}, we showcase the outcomes of federated learning on the same Alpaca dataset with 3 clients and 8 clients, respectively. Initially, there is no significant difference between the two during the early rounds of training. However, after 200 rounds, the training loss with 3 clients has dropped to its lowest and begins to fluctuate, while the training loss with 8 clients continues to steadily converge. This demonstrates that the model converges more stably with more clients participating in the training. This conclusion also corresponds to the theoretical results regarding the number of clients $N$ discussed in Section \ref{sec:main_results}, i.e., an increase in $N$ is beneficial for reducing global convergence.

For a more comprehensive investigation, we further selected client numbers of 3, 5, 9, 20, and 40 on the Alpaca dataset to cover a broad range of scenarios. From the results in Table \ref{table:client_num}, we glean two insights:
\begin{itemize}
\item Initial convergence is quicker with fewer clients, while increasing clients will stabilize convergence. For instance, with the same initial loss, by 200 rounds, the loss of 3-clients is 1.48 compared to 1.51 for 9-clients. And by 400 rounds, the training of 3-clients fails to converge further.
\item Beyond a certain threshold, further increasing client numbers doesn't significantly speed up convergence. When the number of clients exceeds 9, the differences in loss among different clients are less than 1\% throughout 500-round.
\end{itemize}
\begin{table}[h]
\caption{Test loss on a broad range of client number.}\label{table:client_num}
\centering
\setlength\tabcolsep{2pt}

\begin{tabular}{cccccc}
\toprule
\multirow{2}{*}{\textbf{Rounds}}& \multicolumn{5}{c}{\textbf{Test Loss}} \\
\cmidrule(lr){2-6}
& \textbf{3-clients}& \textbf{5-clients}& \textbf{9-clients}& \textbf{20-clients}& \textbf{40-clients} \\

\midrule
\textbf{9}& 2.31& 2.32& 2.31& 2.31& 2.31 \\
\textbf{109}& \textbf{1.94}& 1.95& 1.96& 1.97& 1.98 \\
\textbf{209}& \textbf{1.48}& 1.49& 1.51& 1.50& 1.52 \\
\textbf{309}& 1.51& 1.39& \textbf{1.37}& 1.38& 1.38 \\
\textbf{409}& -& 1.39& \textbf{1.33}& 1.35& 1.34 \\
\textbf{500}& -& 1.47& \textbf{1.31}& 1.33& 1.32 \\
\bottomrule

\end{tabular}
\end{table}

\subsection{The Impact of Batch Size}
\label{appendix:exp_impact_batch_size}

We present the results of altering the batch size on the Dolly-Meta dataset as an example in Table \ref{table:batch_size}. The results show that larger batch-size start with lower loss. For the first 200 epochs, the loss for batch-size=1 is smaller than for the other two. However, larger batch-size decline more slowly, and by the end, batch-size=1 has the smallest test loss. This implies that larger batch-size are unnecessary during training.
\begin{table}[h]
\caption{Effects of various batch sizes on Dolly-Meta dataset.}\label{table:batch_size}
\centering
\setlength\tabcolsep{2pt}
\begin{tabular}{cccc}
\toprule
\multirow{2}{*}{\textbf{Rounds}}& \multicolumn{3}{c}{\textbf{Test Loss}} \\
\cmidrule(lr){2-4}
& \textbf{Batch Size=1}& \textbf{Batch Size=3}& \textbf{Batch Size=5} \\

\midrule
\textbf{9}& 1.87& 1.61& \textbf{1.58} \\
\textbf{109}& 1.68& 1.57& \textbf{1.55} \\
\textbf{209}& \textbf{1.52}& 1.53& 1.53 \\
\textbf{309}& \textbf{1.40}& 1.50& 1.51 \\
\textbf{409}& \textbf{1.38}& 1.47& 1.49 \\
\textbf{500}& -& \textbf{1.43}& 1.47 \\
\bottomrule

\end{tabular}
\end{table}

\subsection{The Impact of Model Size}
\label{appendix:exp_impact_model_size}

To investigate the and versatility of \ours across different model sizes, based on LLaMA2-7B~\cite{touvron2023llama2}, we conducted experiments using the same experimental setup on three datasets: Dolly-Meta, CodeAlpaca-LDA and GSM8K-IID. The experimental results are shown in Table \ref{table:model_size}. The results show that \ours on LLaMA2-7B retains similar trends, while starts with lower Loss than 3B-model by 0.2 and 0.12 in Dolly-Meta and GSM8K-IID. For all tasks, 7B-model's loss decreases more slowly, with reductions at 300 rounds being only 34\%, 67\%, and 41\% of the 3B-model's in Dolly-Meta, Code-LDA, and GSM8K-IID.

\begin{table}[h]
\caption{Test loss on LLaMA-3B and LLaMA2-7B.}\label{table:model_size}
\centering
\setlength\tabcolsep{2pt}
\begin{tabular}{ccccccc}
\toprule
\multirow{2}{*}{\textbf{Rounds}}& \multicolumn{2}{c}{\textbf{Dolly-Meta}}& \multicolumn{2}{c}{\textbf{Code-LDA}}& \multicolumn{2}{c}{\textbf{GSM8K-IID}}\\
\cmidrule(lr){2-3} \cmidrule(lr){4-5} \cmidrule(lr){6-7}
& \textbf{3B}& \textbf{7B}& \textbf{3B}& \textbf{7B}& \textbf{3B}& \textbf{7B} \\

\midrule
\textbf{9}& 1.87& 1.67& 0.93& 0.94& 1.26& 1.14 \\
\textbf{109}& 1.68& 1.59& 0.91& 0.93& 1.10& 1.09 \\
\textbf{209}& 1.52& 1.55& 0.87& 0.91& 1.04& 1.05 \\
\textbf{309}& 1.40& 1.51& 0.87& 0.90& 0.94& 1.01 \\
\bottomrule

\end{tabular}
\end{table}

\subsection{The Impact of Perturbation Scale}
\label{appendix:exp_impact_mu}

\begin{figure*}[h]
    \centering
    \includegraphics[width=0.9\textwidth]{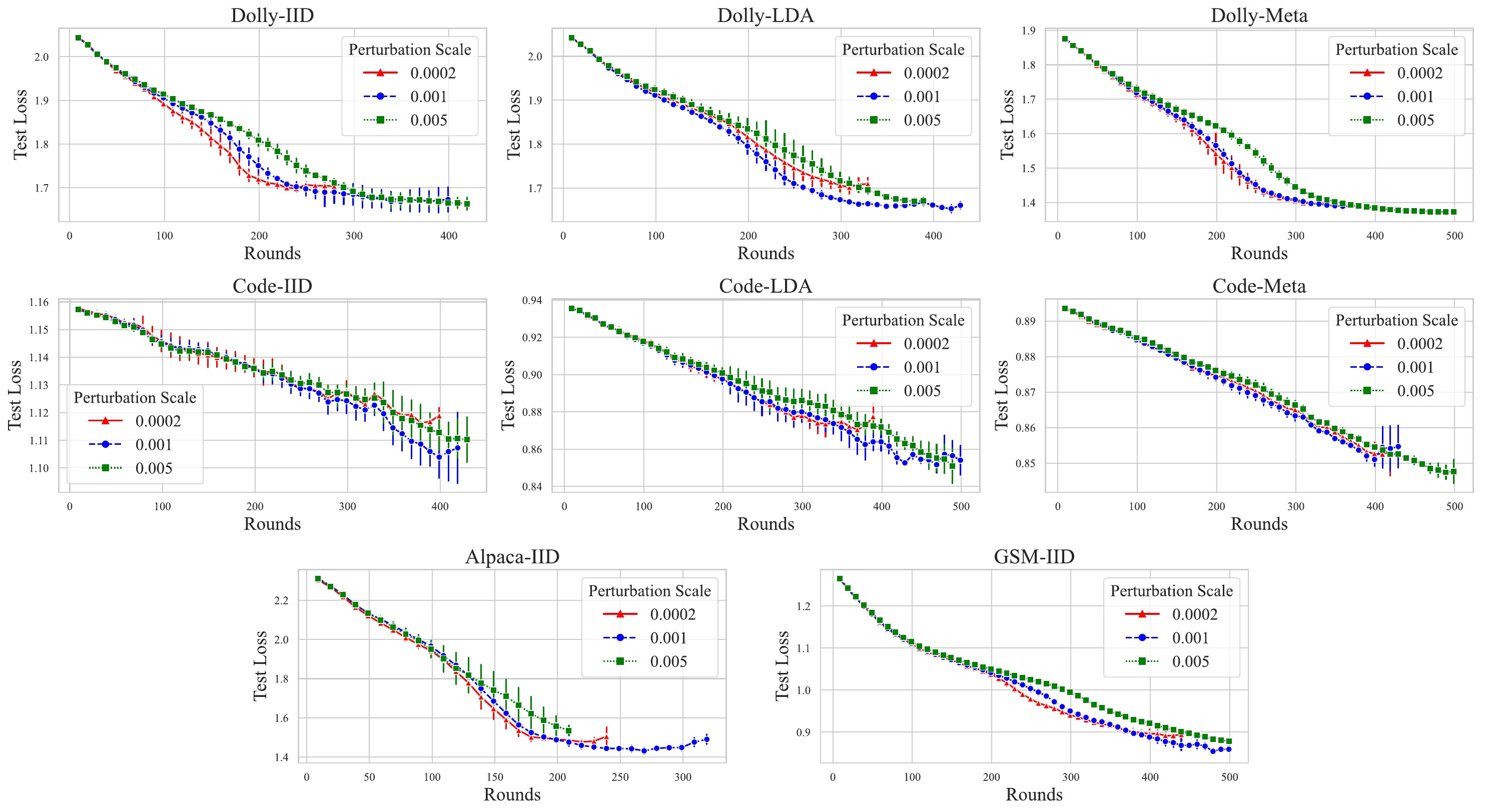}
    \caption{Effects of different perturbation scales $\mu$.}
    \label{fig:diff_mu_app}
\end{figure*}

We display the comprehensive experimental results of altering the perturbation scale $\mu$ across different datasets in Figure \ref{fig:diff_mu_app}. As mentioned in Section \ref{subsubsec:exp_impact_mu}, we observe that, except for CodeAlpaca, smaller values of $\mu$ slightly accelerate the convergence speed in the remaining results, with their corresponding lines all positioned below the default setting, whereas larger values of $\mu$ result in slower convergence speeds. Through these extensive experiments, we further substantiate the theoretical findings.

\subsection{The Impact of Local Iterations}
\label{appendix:exp_impact_H}

We present the complete experimental results of changing the local iteration $H$ on different datasets in Figure \ref{fig:diff_H_app}. As mentioned in Section \ref{subsubsec:exp_impact_h}, we find that across all datasets and splitters, a larger $H$ significantly speeds up convergence, although in certain scenarios, such as Code-IID and Code-Meta, the loss does not reach a lower stable convergence state, and a smaller $H$ consistently results in slower convergence speeds. Through these comprehensive experiments, we further validate the theoretical results in Section \ref{sec:main_results} related to the impact of $H$.

\subsection{Comprehensive Results of Convergence Study}
\label{appendix:exp_main_convergence_app}

\begin{figure*}[h]
    \centering
    \includegraphics[width=0.9\textwidth]{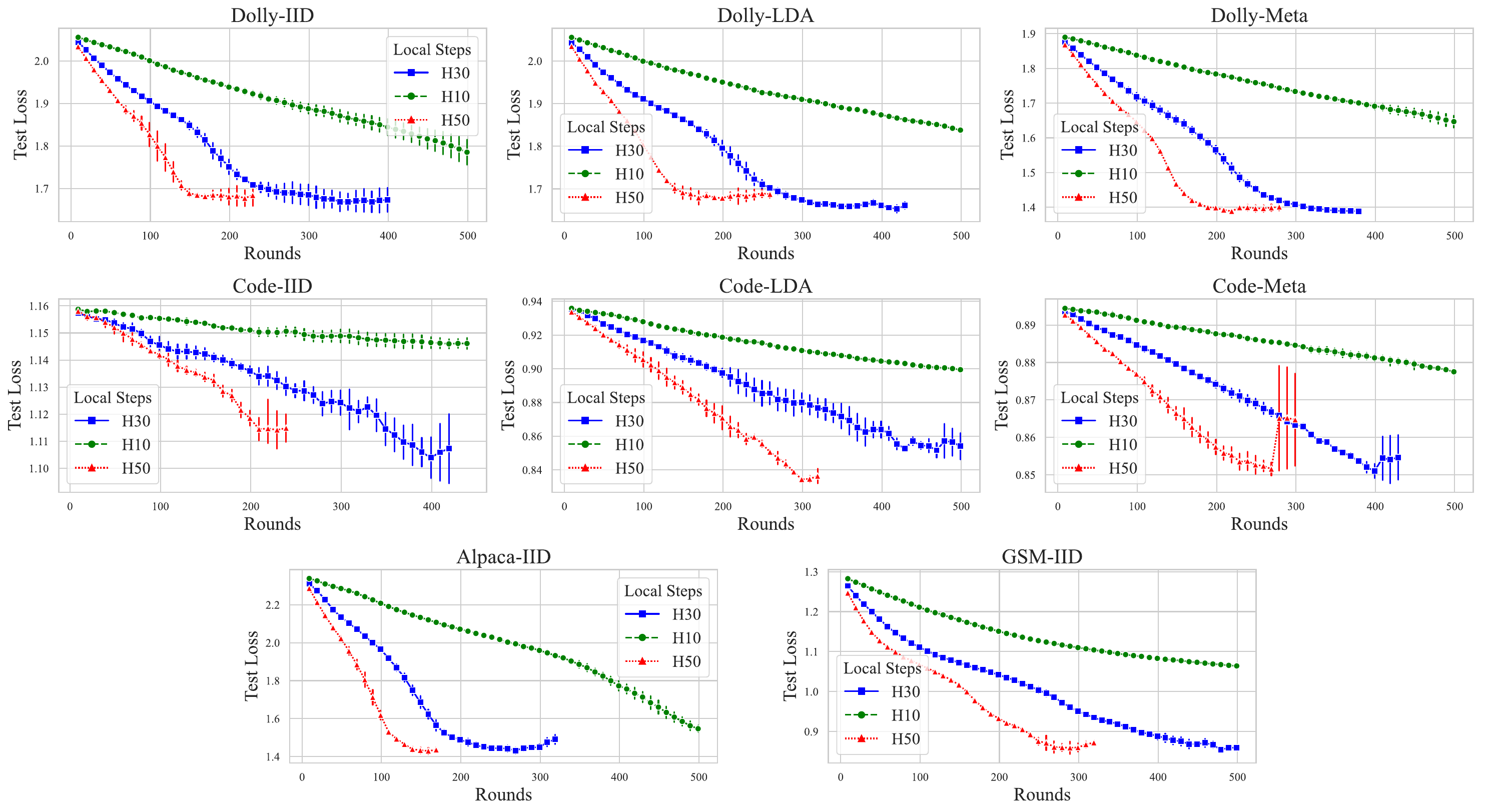}
    \caption{Effects of different local iterations $H$.}
    \label{fig:diff_H_app}
\end{figure*}

\begin{figure*}[h]
    \centering
    \includegraphics[width=0.9\textwidth]{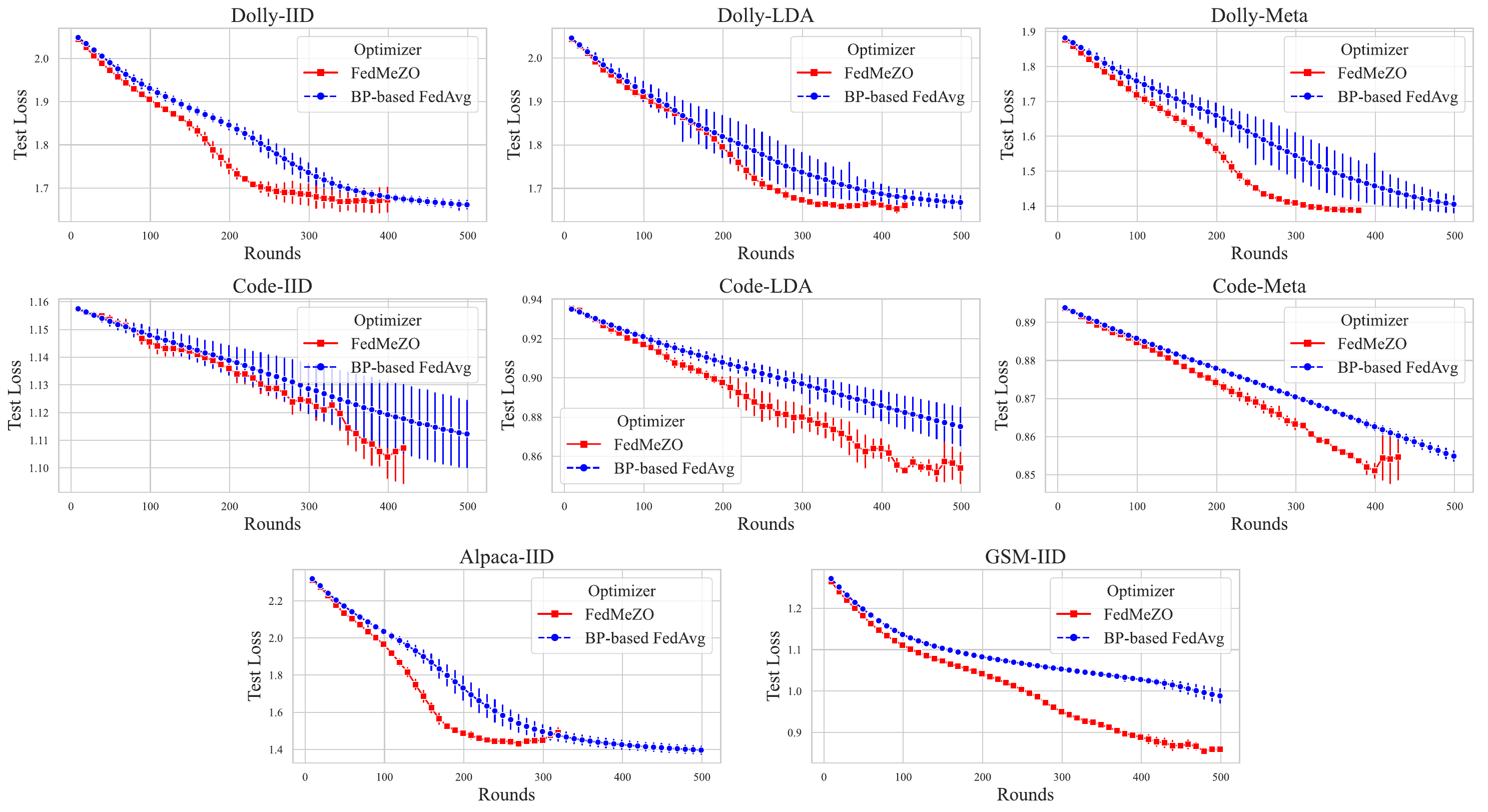}
    \caption{Convergence comparison of \ours and BP-based FedAvg algorithm.}
    \label{fig:main_convergence_app}
\end{figure*}

As mentioned in Section \ref{subsec:exp_main_convergence}, we have validated the convergence of BP-based FedAvg and FedMeZO across different datasets and splitters, with the results displayed in Figure \ref{fig:main_convergence_app}. We observed that in all scenarios, FedMeZO achieves a faster loss reduction compared to BP-based FedAvg and attains a lower loss in certain datasets, such as Code-IID and GSM-IID. This indicates that under equivalent conditions, FedMeZO is more adept at learning the characteristics of different datasets. 
In the CodeAlpaca dataset, we observe that the BP-based FedAvg exhibits fluctuations with large error bars, showing a trend similar to that observed at the end of Code-Meta in Figure \ref{fig:diff_H_app}. 
We hypothesize that the increased fluctuation is attributable to the code data's inherently disjointed nature in natural language terms, compared to datasets from other domains.

\end{document}